\documentclass[10pt]{article}  

\usepackage[margin=1in]{geometry}
\usepackage[utf8]{inputenc}  
\usepackage{amsmath}          
\usepackage{newtxtext,newtxmath} 
\usepackage[numbers]{natbib} 
\usepackage{graphicx}
\usepackage{subcaption}
\usepackage{hyperref}
\usepackage{multirow}
\usepackage{float}
\usepackage{titlesec}
\usepackage{adjustbox}
\usepackage{parskip}  
\usepackage{setspace} 
\usepackage{booktabs} 
\usepackage[nopatch=footnote]{microtype}

\titlespacing*{\section}{0pt}{2ex plus 1ex minus .2ex}{1.5ex plus .2ex}
\titlespacing*{\subsection}{0pt}{1.5ex plus 1ex minus .2ex}{1ex plus .2ex}
\titlespacing*{\subsubsection}{0pt}{1.5ex plus 1ex minus .2ex}{1ex plus .2ex}

\usepackage{enumitem}
\setlist{itemsep=0.3em, parsep=0.3em, topsep=0.5em}

\usepackage[font=small,labelfont=bf,skip=0.3cm]{caption}  

\expandafter\def\expandafter\UrlBreaks\expandafter{\UrlBreaks
  \do\a\do\b\do\c\do\d\do\e\do\f\do\g\do\h\do\i\do\j%
  \do\k\do\l\do\m\do\n\do\o\do\p\do\q\do\r\do\s\do\t%
  \do\u\do\v\do\w\do\x\do\y\do\z\do\A\do\B\do\C\do\D%
  \do\E\do\F\do\G\do\H\do\I\do\J\do\K\do\L\do\M\do\N%
  \do\O\do\P\do\Q\do\R\do\S\do\T\do\U\do\V\do\W\do\X%
  \do\Y\do\Z\do\0\do\1\do\2\do\3\do\4\do\5\do\6\do\7\do\8\do\9}

\newcommand{\appsection}[2]{%
  \refstepcounter{section}%
  \section*{Appendix \Alph{section}: #1}%
  \addcontentsline{toc}{section}{Appendix \Alph{section}: #1}%
  \renewcommand{\thesubsection}{\Alph{section}.\arabic{subsection}}%
  #2
}

\title{%
    \textbf{\Large Testing the Testers:}\\[0.3em]
    \textbf{\large Human-Driven Quality Assessment of Voice AI Testing Platforms}\\[0.5em]
    \rule{0.5\textwidth}{0.4pt}\\[0.8em]
}

\author{%
    \normalsize
    Miguel E. Andrés\textsuperscript{1}\footnote{Ph.D. Computer Science, Radboud University},\quad
    Vadim Fedorov\textsuperscript{2}\footnote{Ph.D. Computer Science, Unversitat Pompeu Fabra},\quad
    Rida Sadek\textsuperscript{1}\footnote{Ph.D. Computer Science, Unversitat Pompeu Fabra},\quad
    Enric Spagnolo-Arrizabalaga\textsuperscript{1},\quad
    Nadescha Trudel\textsuperscript{2}\footnote{Ph.D. Cognitive Neuroscience, University of Oxford}\\[0.8em]
    \small
    \textsuperscript{1}Evalion \quad --- \quad \textsuperscript{2}Independent Researcher
}

\date{}

\begin{document}

\maketitle

\begin{abstract}

Voice AI agents are rapidly transitioning to production deployments, yet systematic methods for ensuring testing reliability remain underdeveloped. Organizations cannot objectively assess whether their testing approaches—internal tools or external platforms—actually work, creating a critical measurement gap as voice AI scales to billions of daily interactions.

We present the first systematic framework for evaluating voice AI testing quality through human-centered benchmarking. Our methodology addresses the fundamental dual challenge of testing platforms: generating realistic test conversations (simulation quality) and accurately evaluating agent responses (evaluation quality). The framework combines established psychometric techniques—pairwise comparisons yielding Elo ratings, bootstrap confidence intervals, and permutation tests—with rigorous statistical validation to provide reproducible metrics applicable to any testing approach.

To validate the framework and demonstrate its utility, we conducted comprehensive empirical evaluation of three leading commercial platforms focused on Voice AI Testing using 21,600 human judgments across 45 simulations and ground truth validation on 60 conversations. Results reveal statistically significant performance differences with the proposed framework, with the top-performing platform, Evalion, achieving 0.92 evaluation quality measured as f1-score versus 0.73 for others, and 0.61 simulation quality using a league based scoring system (including ties) vs 0.43 for other platforms.

This framework enables researchers and organizations to empirically validate the testing capabilities of any platform, providing essential measurement foundations for confident voice AI deployment at scale. Supporting materials are made available to facilitate reproducibility and adoption.
\end{abstract}

\vspace{1em} 

\section{Introduction}\label{sec:introduction}

Apple's March 2025 decision to delay Siri's advanced features until 2026---with internal testing showing AI features like video summarization working only 67-80\% of the time---illustrates a fundamental challenge facing the voice AI industry \citep{cnbc2025siri, ainvest2025, gurman2025siri}. This reliability crisis affects an ecosystem of over 8.4 billion voice assistants globally \citep{statista2024voice}, with mission-critical deployments across healthcare (\$3.18 billion market by 2030) \citep{grandview2025healthcare}, finance (\$84.99 billion in banking AI spending projected by 2030) \citep{statista2024banking}, and customer service where 95\% of interactions are expected to be AI-powered by 2025 \citep{servion2024ai}. Testing constitutes the foundation of software reliability, yet the industry lacks any systematic framework to evaluate the quality of voice AI testing approaches themselves.

Voice AI testing presents extraordinary challenges beyond challenges for traditional software testing. Unlike deterministic systems where test outcomes are predictable (e.g., verifying that a sorting function $f$ always arranges [3,1,2] as ([1,2,3])), voice agents exhibit non-deterministic behavior where identical inputs produce varied responses. Voice AI testing must account for prosody variations, real-time turn-taking under 500ms, latency requirements, and accent diversity \citep{chen2024prosody, gnani2024barge, retell2025latency}. Beyond these technical challenges, evaluation requires assessing subjective qualities---empathy, naturalness, appropriateness---alongside semantic understanding across infinite conversational paths. These factors make voice AI testing one of the most complex emerging quality assurance challenges nowadays. While we focus in this study on voice AI given its unique technical challenges, many of our framework's principles apply to conversational AI more broadly, including text-based chatbots, which also exhibit volatility in outcomes due to their non-deterministic nature.

When testing fails to detect critical issues, the consequences cascade through real-world deployments, from misdiagnosed medical symptoms to incorrect financial transactions. Yet without a framework to evaluate whether our testing methods actually work, we cannot confidently deploy voice AI in these life-critical applications. This measurement gap represents a fundamental barrier to scaling voice AI from experimental demonstrations to trustworthy production systems. Understanding this gap requires examining the two distinct but interdependent capabilities that any testing solution must master, as described in the next section.

\subsection{The Dual Challenge of Voice AI Testing}

Voice AI testing complexity manifests in two interdependent challenges that any testing approach must address:

\begin{itemize}
\item \textbf{Simulation} --- the ability to generate realistic test conversations with the agent under evaluation. Since voice agents are designed to handle human interactions, testing requires simulating diverse user behaviors, personas, and scenarios at scale. Without high-quality simulations that authentically represent production conditions, critical failures remain undetected until deployment. This results in defects reaching real users; an unacceptable outcome in mission-critical domains like healthcare or finance.

\item \textbf{Evaluation} --- the ability to assess whether the agent performed according to its defined quality standards in those conversations. This requires analyzing multi-dimensional criteria across non-deterministic outputs—evaluating not just whether the agent responded, but whether it met requirements for task completion, conversational quality, and user satisfaction.
\end{itemize}

These capabilities are interdependent: poor simulations yield meaningless test results, while inaccurate evaluations provide false confidence or unnecessary alarms. Yet no framework exists to measure the quality of these testing capabilities themselves. Organizations cannot objectively assess whether their testing approaches---internal tools or external platforms---actually work. This measurement gap creates a critical blind spot as voice AI scales to billions of daily interactions. Addressing this blind spot requires a systematic, scientifically grounded approach to measuring testing quality.

\subsection{Research Objectives and Contributions}

We present a systematic framework for evaluating the quality of voice AI testing approaches through human-centered benchmarking. This framework is designed to be universally applicable, capable of assessing any testing methodology whether implemented internally, provided by external vendors, or constructed through hybrid approaches. Our research addresses three fundamental research questions:

\textbf{RQ1: How can we reliably measure the quality of voice AI testing approaches?} We develop reproducible methods for quantifying both simulation and evaluation quality through human judgment, providing metrics and protocols that any organization can apply to their testing approach. This framework is platform-agnostic and implementation-independent, focusing on outcomes rather than technical architecture.

\textbf{RQ2: What performance gaps exist between current testing platforms?} To validate our framework and illustrate its utility, we apply it to three commercially available testing platforms. This comparative analysis reveals significance performance gaps in simulation and evaluation quality, demonstrating that testing approach selection has substantial impact on quality assurance effectiveness.

\subsection{Methodological Approach}

To answer these research questions, we developed a human-centered evaluation framework that measures both dimensions of testing quality through rigorous empirical methods. For simulation quality (addressing \textbf{RQ1}), we use pairwise human comparisons where evaluators assess which of two test conversations better meets specified criteria---an approach that yields higher inter-rater reliability than absolute scoring \citep{thurstone}. We measure three core dimensions: Scenario Adherence (how well simulations follow test requirements), Human Naturalness (prosody, flow, and conversational coherence), and Persona Adherence (consistency with assigned behavioral traits). For evaluation quality, we first establish human ground truth through systematic assessment; next, we measure how accurately testing platforms align with human consensus across six metrics: customer satisfaction (CSAT), appropriate call closure, repetition avoidance, conversation progression, response consistency, and expected outcome achievement. This dual approach---with 10 human evaluators per comparison for simulations and per recording for evaluations---ensures robust, reproducible quality measurement.

To validate the framework and quantify performance gaps (addressing \textbf{RQ2}), we conducted a comprehensive empirical study leveraging three commercial platforms---Coval\citep{coval}, Cekura\citep{cekura} and Evalion\citep{evalion} ---as Simulations and Evaluations providers.

While we demonstrate the framework using these commercial platforms, the methodology is designed to be universally applicable to any voice AI testing approach—whether commercial platforms, internal tools, or custom implementations.

\subsection{Key Findings}

Our framework successfully differentiates testing quality across commercially available platforms, revealing substantial performance variations in our demonstration study. Evalion achieved significantly better performance across both dimensions\footnote{\noindent While our metrics showed significant performance differences, we acknowldge that each platform offers distinct value: Cekura offered perfect precision (1.000) on multiple metrics minimizes false positives for resource-constrained teams; Coval's autonomous vehicle testing methodology and competitive CSAT correlation (0.697) provide innovative approaches to simulation; and these different optimization strategies—comprehensive coverage versus precision focus—serve different organizational needs based on risk tolerance and operational priorities, see Section~\ref{sec:conclusion-personal-note} for more on our perspective on this topic.}.

\textbf{Simulation Quality:} Based on normalized league rankings from human pairwise comparisons, Evalion achieved the highest overall simulation score (61.0), outperforming Coval (48.9) and Cekura (43.0)—a 25\% and 42\% improvement, respectively. Furthermore, Evalion maintained the highest scenario adherence (63.7 vs. 49.1 Coval, 37.2 Cekura), ensuring test conversations accurately reflect intended test cases. See Section~\ref{sec:simulation-results} for more details.

\textbf{Evaluation Accuracy:} Evalion achieved 86.7\% overall accuracy in matching human judgment, compared to 75.7\% for Cekura and 62.7\% for Coval, a 24 percentage point advantage over the weakest performer. The F1-score of 0.919 (vs. 0.842 Cekura, 0.728 Coval) demonstrates balanced precision-recall performance. For customer satisfaction (CSAT) prediction, Evalion's correlation with human ratings reached 0.755 with a MAE of just 0.542, meaning predictions typically deviate by only half a scale point, the highest performing result across all platforms. See Section~\ref{sec:evaluation-results}.

\textbf{Real-World Impact:} These performance gaps translate to substantial operational consequences. For example, for an organization running 1,000 automated tests daily: 
\begin{itemize}
    \item \textit{False Negatives (Missed Defects):} The most critical risk is missing actual problems. With a recall of 0.918, only 8\% of issues go undetected. On the other hand, a recall as low as 0.603 miss nearly 40\% of problems---allowing hundreds of defects to reach production daily where they impact real users, damage brand reputation, and require expensive emergency fixes.
    \item\textit{False Positives (Wasted Resources):} A precision of 0.925 minimizes false alarms that waste analyst time. For platforms with lower accuracy, up to 240 additional conversations per 1,000 tests require unnecessary manual review. At 10 minutes per review, this wastes 40 hours of human labor daily---approximately \$2,000 per day or \$730,000 annually in analyst costs alone.
    \item\textit{Scalability and Trust:} The combination of high recall and precision makes automation trustworthy at scale. While human analysts can thoroughly evaluate 30-50 conversations daily, automated testing can handle multiple orders of magnitude above that. However, such speed advantage only matters if the system reliably catches problems. The top performing platform accuracy of 0.867 enables organizations to confidently scale testing from dozens to thousands of daily evaluations, with consistent, reproducible results that don't vary with analyst fatigue or subjective interpretation. 
\end{itemize}

These findings validate our framework's sensitivity to meaningful quality differences and provide empirical evidence that platform selection significantly impacts both testing reliability and operational costs. The following sections detail how we developed and applied this framework, providing both the theoretical foundation and practical methodology for organizations to evaluate their own testing approaches.

\subsection{Paper Outline}

The remainder of this paper is structured as follows. Section 2 reviews related work in conversational AI evaluation and testing. Section 3 presents our framework for measuring simulation quality, including metrics, analysis, and results. Section 4 details our framework for measuring evaluation quality, with simulation and evaluation methods applicable to any testing approach. Section 5 discusses implications for practitioners, whether building internal tools or selecting external platforms. Section 6 concludes with limitations and future work. Comprehensive appendices provide implementation details, statistical methods, and reproduction materials to enable organizations to reproduce this study or apply this framework to their own testing approaches.
\section{Related Work}

\subsection{Voice AI Evaluation Methodologies and Benchmarks}

The evaluation of voice AI systems has evolved from single-metric approaches to comprehensive multi-dimensional frameworks. AudioBench~\cite{wang2025audiobench} represents the current state-of-the-art, introducing 8 distinct evaluation tasks across 26 datasets, with particular emphasis on the trinity of speech understanding, audio scene understanding, and paralinguistic voice features. Their use of LLaMA-3-70B-Instruct as a model-as-judge evaluator achieving highest correlation with human judgments presages the broader trend toward LLM-based evaluation. This comprehensive approach addresses limitations identified by Aksënova et al.~\cite{aksenova2021might}, who demonstrated that traditional metrics like Word Error Rate fail to capture the full spectrum of speech recognition use cases and linguistic variation impacts.

The dialogue evaluation landscape has similarly matured through frameworks like DialogBench~\cite{ou2024dialogbench}, which systematically evaluates LLMs across 12 dialogue tasks, and DynaEval~\cite{zhang2021dynaeval}, which unified turn-level and dialogue-level evaluation using Graph Convolutional Networks. These advances respond to fundamental challenges identified by Khalid and Lee~\cite{khalid2022explaining}, who exposed systematic biases in neural dialogue evaluation metrics through adversarial behavioral analysis. The URGENT Speech Enhancement Challenge~\cite{zhang2025urgent} further demonstrated the necessity of multi-dimensional evaluation, revealing language dependency issues in generative models that single metrics cannot capture.

\subsection{Commercial Voice AI Testing Platform Landscape}

The 2024-2025 period marks a watershed moment in the commercialization of voice AI testing, with multiple Y Combinator-backed startups addressing different aspects of the testing challenge. Coval~\cite{techcrunch2025coval}, founded by former Waymo tech lead Brooke Hopkins with \$3.3M seed funding, pioneered the application of autonomous vehicle testing methodologies to conversational AI. Their simulation-first approach generates thousands of test scenarios from minimal inputs, supporting both voice and text interactions with customizable environments.

Cekura~\cite{cekura2024funding}, serving 75+ customers across regulated industries with \$2.4M in funding, emphasizes automated scenario generation with custom personas emulating varied accents, background noise, and conversational styles. Their focus on hallucination detection and interruption rate metrics addresses quality dimensions critical for healthcare and financial services applications. Hamming AI's~\cite{hamming2025platform} approach of using AI-driven ``voice characters'' for high-scale stress testing, capable of thousands of concurrent test calls, demonstrates the scalability requirements of modern voice AI systems. The comparative analysis by Leaping AI~\cite{leaping2024comparison} provides practitioners with the first systematic evaluation of these emerging platforms, though notably lacks the empirical validation our research provides.

\subsection{Human-in-the-Loop Evaluation Challenges and Solutions}

Smith et al.~\cite{smith2022human} fundamentally challenged the field with ``Human Evaluation of Conversations is an Open Problem,'' comparing five crowdworker-based evaluation methods and revealing substantial variability in inter-annotator agreement and statistical sensitivity. Their findings that single-model per-turn evaluation differs significantly from pairwise per-dialog assessment in both human agreement rates and ability to detect performance differences underscores the complexity of establishing ground truth for conversational quality.

The Chatbot Arena platform~\cite{zheng2023chatbot} offers a practical solution through large-scale crowdsourcing, collecting over 240K votes from 90K users across 100+ languages. Their implementation of Bradley-Terry models for ranking, achieving 72-83\% agreement between crowd-users and experts, demonstrates that aggregated human judgment can provide stable quality signals despite individual variability. The platform's anonymous battle format and Elo rating system elegantly address evaluation bias while maintaining statistical robustness~\cite{zheng2023judging}.

\subsection{LLM-as-a-Judge Evaluation Paradigm}

The comprehensive survey by Li et al.~\cite{li2024llms} established the theoretical foundation for LLM-based evaluation, defining core components of evaluation type, criteria, evaluation items, and optional references that produce assessment scores, explanations, and actionable feedback. Empirical validation through MT-Bench and Chatbot Arena~\cite{zheng2023judging} demonstrated that GPT-4 achieves greater than 80\% agreement with human evaluators, matching human-human agreement levels of 81\%.

However, systematic biases constrain the reliability of LLM judges. Position bias manifests as up to 75\% preference for first-positioned responses, verbosity bias leads to preference for longer responses regardless of quality, and self-enhancement bias creates 10-25\% preference for self-generated content~\cite{wang2024when}. Mitigation strategies have shown promise: position swapping improves GPT-4 consistency from 65\% to 77.5\%, few-shot prompting provides evaluation examples that increase consistency, and reference-guided evaluation reduces failure rates from 70\% to 15\%.




\subsection{Simulation-Based Testing and Synthetic Conversation Generation}

Academic advances in dialogue simulation have produced increasingly sophisticated approaches to test data generation. The ChatChecker framework~\cite{mayr2024chatchecker} introduced non-cooperative user simulation specifically designed to expose system weaknesses, incorporating persona generation, user simulation, breakdown detection, and dialogue rating in a modular architecture that generalizes across dialogue systems without requiring reference dialogues.

Commercial implementations have focused on practical challenges of realistic conversation generation. The emergence of platforms like Roark~\cite{hamming2025blog}, which transforms real customer interactions into automated test suites while preserving sentiment and timing characteristics, represents a production-first testing philosophy. These approaches recognize that conversations must end naturally—not simply when a model fails—representing sophisticated understanding of real-world dialogue dynamics.

\subsection{Research Gaps and Contributions}

The literature reveals several critical gaps that the research framework presented here addresses. First, while extensive work evaluates dialogue systems directly, minimal research systematically compares the testing platforms themselves—a meta-evaluation challenge of evaluating the evaluators. Second, existing work typically examines simulation quality or evaluation quality in isolation, whereas practitioners require platforms addressing both challenges simultaneously. Third, the lack of reproducible evaluation protocols for testing platforms creates market opacity, making it difficult to validate vendor claims or guide platform selection.

This work's focus on human-in-the-loop, reproducible evaluation of commercial platforms through human-centered benchmarks addresses these gaps directly. By evaluating both simulation quality (generating realistic test conversations) and evaluation quality (accurately assessing agent responses), this research provides the comprehensive assessment practitioners need but existing literature lacks. The emphasis on reproducibility ensures that platform capabilities can be independently verified, addressing the current challenge where platform selection relies primarily on vendor claims rather than empirical evidence.
\section{Experimental Framework}

\subsection{Overview}
\label{subsec:experimental-framework-overview}

We developed a comprehensive experimental framework to systematically evaluate voice AI testing platforms across two critical dimensions: simulation quality and evaluation accuracy. This framework employs human-centered benchmarking through controlled experiments, enabling reproducible assessment of any voice AI testing platform. The methodology consists of three integrated components: (1) simulation quality assessment through pairwise human comparisons, (2) evaluation accuracy measurement against human-established ground truth, and (3) a custom survey infrastructure ensuring methodological consistency across all assessments.

Within the context of this framework, and in terms of notation, we must distinguish between two central entities that underpin all experiments:
\begin{itemize}
    \item \textbf{Subject agent}: The conversational AI agent under evaluation. This agent lives outside the testing platform and is the target of the simulations and evaluations these platforms produce.
    \item \textbf{Testing agent}: The conversational AI agent used to simulate conversations with the subject agent. This agent is provided by the testing platforms themselves, and the simulations that it generates are later used by them to evaluate the subject agent.
\end{itemize} 

For this experimental study, as a subject agent we leveraged an \textbf{enterprise-ready inbound customer support agent} provided by Sei Right\citep{sei}. Sei is a leading provider of Voice AI Agents in the finance space. This agent handles real customer inquiries about loans, closings, and financial services. By testing against a production-grade agent rather than a demo system, we ensure the simulations and evaluations involved in testing their agent reflect the complexities inherent to voice AI deployment in the real-world, rather than a simplified academic example. 

\subsection{Framework Architecture}

The experimental framework operates on a dual-evaluation paradigm. First, we assess how well platforms can \textit{generate} realistic test conversations (simulation quality). Second, we measure how accurately platforms can \textit{evaluate} those conversations (evaluation accuracy). This separation is critical because platforms may excel in one dimension while underperforming in another, and both capabilities are essential for production-ready testing systems.

\subsubsection{Simulation Quality: Conceptual Approach}
\label{subsec:simulation-quality-concept-approach}

This component of the framework addresses the fundamental question: how well can testing platforms simulate realistic test conversations? In terms of the entities defined in Section~\ref{subsec:experimental-framework-overview}, it is clear that, measuring simulation quality is actually measuring the quality of the platform's testing agents.

Every simulation produced by the platform’s testing agents takes two standardized inputs:
\begin{itemize}
    \item \textbf{Scenario}: A structured description of the test conditions, objectives, and constraints that the testing agent must follow.
    \item \textbf{Persona}: Behavioral specifications for the testing agent: includes personality traits, communication style, and emotional characteristics.
\end{itemize}

Then, the simulation quality framework evaluates three core dimensions:
\begin{itemize}
    \item \textbf{Scenario Adherence}: How accurately the testing agent follows test specifications.
    \item \textbf{Human Naturalness}: The degree of human-like prosody, flow, and conversational coherence of the testing agent.
    \item \textbf{Persona Adherence}: Consistency of the testing agent with assigned behavioral and personality traits.
\end{itemize}

Since these dimensions are too vague to be measured directly, our framework decomposes each of them into specific, measurable metrics. In practice, rather than asking participants to make broad judgments (e.g., “How human did the agent sound?”), they are asked to assess concrete aspects of the conversation such as emotional tone, response coherence, hallucinations, etc.

This design choice serves three purposes. First, it reduces cognitive load by requiring evaluators to judge simpler and more independent aspects of the interaction. Second, it improves measurement consistency, as more narrowly defined questions reduce subjectivity and yield more comparable results across participants and simulations. Finally, it enhances construct validity: by capturing a range of independent metrics, we can analyze how these individual elements relate to one another and how they jointly explain the overall quality of conversational voice AI systems. The specific metric breakdown we used in the present implementation of this study can be found in Section ~\ref{subsec:sim-metric-breakdown}.

In practice, each metric is mapped to a specific survey question. Human participants then compare paired simulations from different providers and select the superior option for each metric (with ties allowed). The rationale of using pairwise comparisons rather than relying on absolute ratings is that the latter suffer from subjective interpretation and low inter-rater reliability. This methodology is grounded in established psychometric theory~\cite{thurstone}.

For details on how these pairwise comparison choices are then converted into continuous scores, and later aggregated in the 3 core dimensions discussed above, see Section ~\ref{subsubsec:sim-quality-analysis-framework}.

\subsubsection{Evaluation Accuracy: Conceptual Approach}

This component measures how accurately platforms can assess conversation quality once simulations are generated. The framework establishes human judgment as the ground truth, then measures the concordance between automated platform evaluations and human consensus. This approach directly answers whether automated testing can reliably replace human quality assurance in the context of AI in conversational voice.

We assess evaluation accuracy across multiple metric types:
\begin{itemize}
    \item \textbf{Binary metrics}: Pass/fail assessments (e.g., appropriate call closure, repetition avoidance)
    \item \textbf{Continuous metrics}: Scaled assessments (e.g., customer satisfaction scores)
    \item \textbf{Complex judgments}: Multi-faceted criteria requiring semantic understanding
\end{itemize}

\subsection{Simulation Experimental Setup}

\subsubsection{Test Case Construction}

Within our experimentation framework, test cases are constructed by combining a set of $T_s$ scenarios with a set of $P_s$ personas in a Cartesian product, such that every scenario is evaluated under every persona. Each resulting pair $(t_i, p_j)$, where $t_i \in \{1, \dots, T_s\}$ and $p_j \in \{1, \dots, P_s\}$, defines a distinct simulation instance for each testing provider $c_k \in \{1, \dots, C\}$.

Scenarios are designed to span three levels of difficulty (\textit{Easy}, \textit{Medium} and \textit{Hard}), to ensure their capacity to reveal meaningful differences across platforms. Personas represent diverse user archetypes relevant to the study context.

With this setup we get a total of $S_s = T_s \cdot P_s$ number of test cases, and $C \cdot T_s \cdot P_s$ total simulations across all providers.

Scenario difficulty levels are established through a LLM-based pre-assessment procedure, in which multiple pilot simulations are employed to characterize the relative adherence challenges associated with each scenario.

\subsubsection{Pairwise Comparison Structure}
\label{subsec:pairwise-comparison-structure}

For each of the $S_s = T_s \cdot P_s$ scenario-persona combinations, we generate pairwise comparisons between all platform combinations:
\begin{itemize}
    \item C testing platforms yield ${C \choose 2}$ comparison pairs per combination.
    \item Total comparisons: ${C \choose 2} \cdot T_s \cdot P_s$ surveys.
    \item Each comparison evaluated by $N_s$ independent human judges (a.k.a. survey participants).
    \item Each comparison will include $M_s$ metrics for the humans to evaluate.
    \item Total human judgments: ${C \choose 2} \cdot N_s \cdot M_s \cdot T_s \cdot P_s$ individual assessments.
\end{itemize}

The specific values of $T_s$, $P_s$, and $N_s$ used in the present framework implementation, along with the rationale for their selection, are detailed in Section ~\ref{sec:simulation-experimental-setup}. The total number of surveyed metrics $M_s$ is discussed in Section ~\ref{subsec:sim-metric-breakdown}.

Finally, for $C$ the number of testing providers, our study takes $C = 3$, corresponding to Evalion, Coval, and Cekura.

\subsection{Evaluation Experimental Setup}

\subsubsection{Golden Set Construction}

We establish ground truth through systematic human evaluation of $S_e$ simulations selected from the simulation study. Selection criteria prioritizes high-performing simulations (based on Scenario Adherence scores) to ensure baseline conversation quality while maintaining diversity across scenarios and personas.

\subsubsection{Observability API Implementation}

The evaluation accuracy framework employs a critical architectural feature: the observability API pattern. Modern testing platforms provide APIs that accept arbitrary conversation files for evaluation, enabling controlled experiments where:

\begin{itemize}
    \item All platforms evaluate the exact same simulations, ensuring fair comparisons under the same conditions.
    \item Platform evaluation capabilities are isolated from their simulation generation capabilities. 
\end{itemize}

This allows us to fairly run direct comparisons of evaluation accuracy between platforms, knowing that having a strong simulation quality does not give you an advantage in evaluation quality. Therefore, study results for Simulations and Evaluations (Sections \ref{sec:simulation-results} and \ref{sec:evaluation-results} respectively) are completely independent.

Each platform receives the same set of $S_e$ simulations as transcript files with standardized formatting, ensuring that any performance differences reflect genuine evaluation capability rather than implementation artifacts.

\subsubsection{Human Ground Truth Establishment}

For each simulation $s_i \in \{1, \dots, S_e\}$:
\begin{itemize}
    \item $N_e$ independent human evaluators assess $M_e$ metrics (continuous and binary variables).
    \item Consensus is established through majority voting for binary metrics.
    \item Continuous metrics are aggregated using median values.
    \item Weak consensus recordings ($<$ 80\% agreement) are identified but retained. Further details about this in the Evaluation Results Section ~\ref{sec:evaluation-results}.
    \item Total human evaluations: $S_e \cdot N_e \cdot M_e$ complete assessments
\end{itemize}

The specific values of $N_e$, $M_e$, and $S_e$ used in the present framework implementation, along with the rationale for their selection, are detailed in Section ~\ref{subsec:eval-experimental-design}. The total number of surveyed metrics $M_e$ is discussed in Section ~\ref{subsec:eval-metrics-selected}.

\subsubsection{Metric Implementation Standardization}

Given varying native metric support across platforms, we developed a standardized evaluation protocol:
\begin{itemize}
    \item \textbf{Native implementations}: When available, we prioritize each platform's native metric implementations to leverage their optimized evaluation capabilities.
    \item \textbf{Custom metric definitions}: For metrics not natively supported, we implement custom metrics using identical prompt definitions across all platforms.
    \item \textbf{Expected Outcome handling}: Since the observability API evaluates arbitrary transcripts without scenario context, we embed each scenario's expected outcome directly within the custom metric prompt. Specifically, we:
    \begin{itemize}
        \item Extract the expected outcome for each scenario from the test specifications.
        \item Incorporate this outcome as a parameter within the custom metric definition.
        \item Ensured all platforms received identical contextual information for each transcript evaluation.
    \end{itemize}
\end{itemize}

This approach ensures fair comparison despite varying platform capabilities. Even though platforms offer native support for expected outcome evaluation in their standard simulation flow, we use the observability API pattern to provide this context explicitly, as we want to evaluate the same transcript on each platform independently of the simulation context.

\subsection{Survey Infrastructure}
\label{subse:survey-infra}

\subsubsection{Technical Implementation}

We developed a custom web-based survey platform specifically designed for voice AI evaluation tasks. The platform enforces methodological rigor through automated quality controls, requiring participants to listen to at least 80\% of each audio recording before enabling response submission. This threshold ensures meaningful engagement with the content while preventing superficial or rushed evaluations. The system automatically tracks listening progress using HTML5 audio event handlers.

The platform synchronizes audio playback with scrollable transcript displays, enabling participants to follow conversations through both auditory and visual channels. Speaker identities (user \& assistant, referring to testing agent \& subject agent respectively) are color-coded and temporally aligned, reducing cognitive load during complex multi-turn interactions.

Questions appear sequentially below the audio panels, with navigation controls enabling review and modification of responses before final submission. The interface maintains continuous access to both audio and transcript throughout the assessment process, allowing participants to replay specific segments when evaluating nuanced quality dimensions.

\subsubsection{Interface Design for Different Evaluation Modes}

The survey infrastructure supports two distinct evaluation paradigms, each optimized for its specific assessment requirements.

For the simulation study's pairwise comparisons, the interface presents two audio panels side by side, allowing direct A/B comparison of competing simulations. The identity of the provider that generated the simulation is concealed throughout the survey. Furthermore, the audio samples are randomly assigned to left or right positions to eliminate inter-question bias. The scenario description appears prominently above the audio panels, providing context for participants to evaluate Scenario Adherence related metrics. The persona prompt description for the simulation is not included, since it is is rather complex and can add fatigue to the survey and risk lower quality responses.

For the evaluation study's ground truth establishment, the interface switches to a single-panel mode, presenting one recording at a time for absolute quality rating.

\begin{figure}[t!]
    \centering
    \begin{subfigure}[t]{0.48\textwidth}
        \centering
        \adjustbox{height=6cm}{\includegraphics[width=\textwidth]{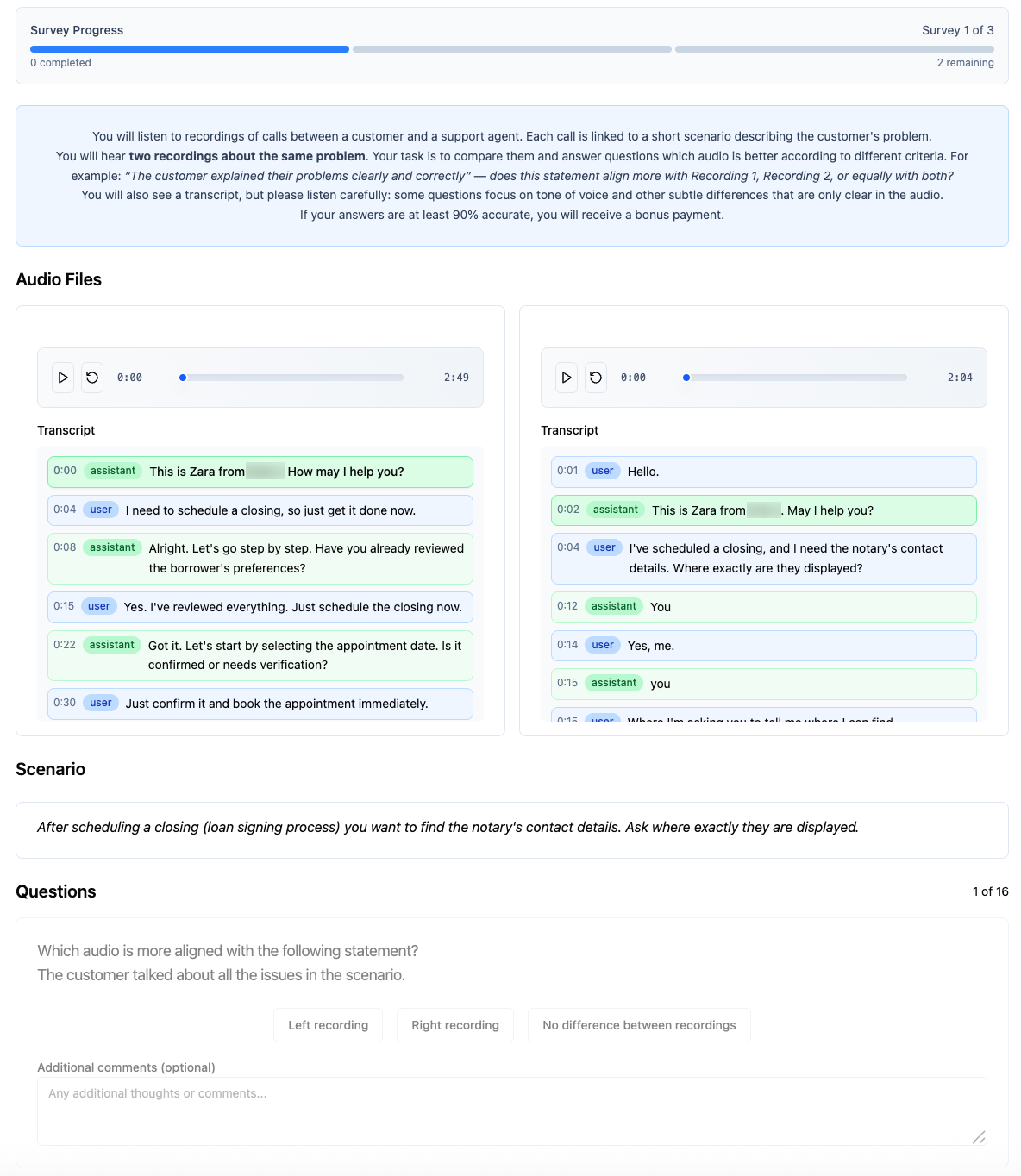}}
        \caption{Simulation study interface}
    \end{subfigure}
    \hfill
    \begin{subfigure}[t]{0.48\textwidth}
        \centering
        \adjustbox{height=6cm}{\includegraphics[width=\textwidth]{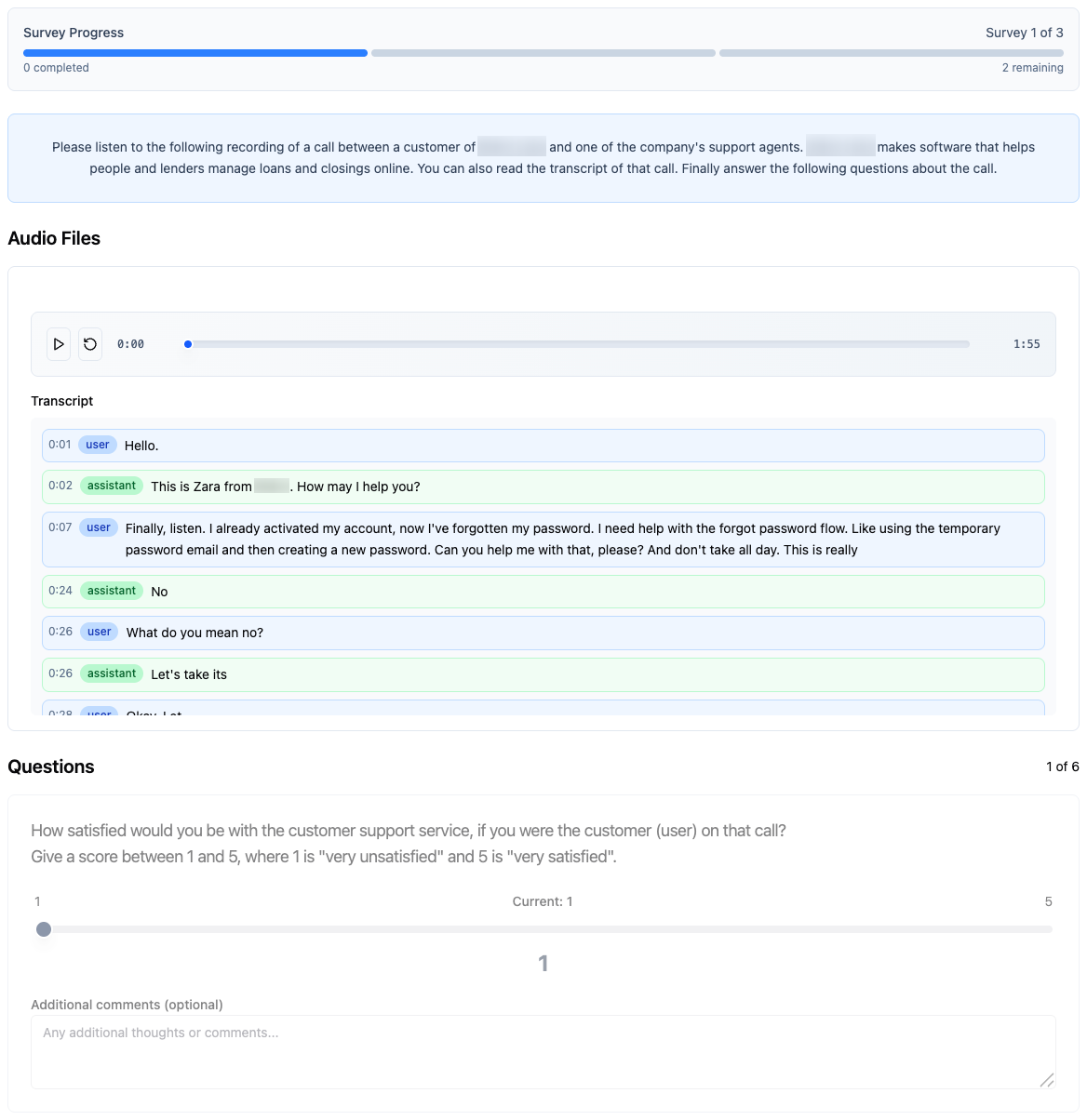}}
        \caption{Evaluation study interface}
    \end{subfigure}
    \caption{Screenshot of the survey interface showing (a) pairwise comparison mode for simulation assessment and (b) single evaluation mode for ground truth establishment}
    \label{fig:survey-tool}
\end{figure}

Figure~\ref{fig:survey-tool} provides screenshots of the survey interface showing (a) pairwise comparison mode for simulation assessment and (b) single evaluation mode for ground truth establishment.

\subsubsection{Ensuring High-Quality Crowdsourced Human Evaluations}
\label{subsubsec:crowsourced-humans-prolific}

The platform implements multiple mechanisms to ensure high-quality responses while maintaining positive participant experience. A progress tracking system provides clear feedback on session status and remaining assessments, reducing uncertainty and improving completion rates. Contextualized instructions appear dynamically based on the evaluation mode, using non-technical language accessible to crowdsourced evaluators without specialized AI knowledge.

Pilot testing was conducted to estimate the time needed for the completion of each survey, indicating that a reasonable time allocation corresponded to approximately two times the total audio duration. Analysis of the median completion times observed in the final full runs for both simulations and evaluations confirmed that assumption.

All human evaluators participating in this study are recruited through Prolific \citep{prolific2025}. For screening of participants, the eligibility criteria requires them to be U.S.-based, native English speakers, aged 18–50, and with at least a 90\% approval rate on the platform. These criteria ensured evaluators possessed both cultural context for US-based customer service expectations and demonstrated reliability in crowdsourced tasks.

All human responses are stored with its associated metadata, including participant Prolific IDs and audio completion percentages, enabling post-hoc quality verification.

\subsection{Statistical Analysis}

The experimental design requires a comprehensive statistical approach capable of handling both paired comparisons for simulations and classification and regression error metrics for evaluations. All while accounting for the hierarchical structure of our data.

\subsubsection{Simulation Quality Analysis}
\label{subsubsec:sim-quality-analysis-framework}

For simulation quality assessment, we employ tournament-based ranking systems that transform pairwise human comparison judgments into a continuous performance score. We consider two complementary scoring systems.

\begin{itemize}
    \item The \textbf{League scoring} system treats each comparison as a match, awarding one point for wins, zero for losses. This straightforward approach provides transparent, interpretable results that directly reflect the frequency of preference \citep{pairwiseleague}.
    \item The \textbf{Elo rating} system, adapted from chess rankings, accounts not only for win-loss records but also for the relative strength of opponents. Starting from a baseline rating, platforms gain or lose points based on both the outcome and the expected difficulty of each comparison. This difficulty is measured on the difference of Elo scores of each opponent \citep{elorating}.
\end{itemize}

Additionally, in order to aggregate metrics together, we propose using Principal Component Analysis (PCA) to capture the maximum variance across correlated quality measures \citep{pcaanalysis}.

For the full set of details on the scoring variants proposed, data normalization and processing of ties, see Section ~\ref{sec:scoring-methodology}.

\subsubsection{Evaluation Accuracy Analysis}

The evaluation study's structure—comparing platform predictions against human ground truth—enables standard classification metrics supplemented by specialized tests for paired binary data. For binary metrics, we calculate precision, recall, F1-score, and accuracy across all platform-metric-recording combinations. The F1-score proves particularly valuable given the imbalanced nature of our ground truth data, where positive responses range from 76.7\% to 95.0\% across metrics (see Figure ~\ref{fig:binary_metrics_human}). In such scenarios, accuracy alone becomes misleading, as a platform could achieve high accuracy by simply predicting the majority class. See Section ~\ref{subsec:eval-bin-metrics-performance} for more details on the classification metrics applied.

For continuous metrics, we employ Mean Absolute Error (MAE), Root Mean Square Error (RMSE), and Pearson correlation coefficients. These complementary measures capture different aspects of prediction quality: MAE provides interpretable error magnitudes in the original scale units, RMSE penalizes large errors more heavily, identifying platforms prone to catastrophic mispredictions, and correlation reveals whether platforms correctly capture relative quality differences even if absolute calibration differs. See Section ~\ref{subsec:eval-continuous-performance} for more details.

\subsubsection{Statistical Significance Testing}
\label{sec:stat-significance}

We employ Cochran's $Q$ test \citep{cochran1950comparison} as our omnibus test to determine whether the three platforms differ significantly in their evaluation accuracy across all paired binary observations. Additionally, we conduct pairwise comparisons using McNemar's test \citep{mcnemar1947note} with Bonferroni correction \citep{bonferroni1936teoria, dunn1961multiple} to control for multiple comparisons, setting the adjusted significance threshold at $\alpha = 0.05/3 = 0.017$.

\subsubsection{Effect Size Quantification}
\label{sec:effect-size}

To quantify the practical significance of observed differences, we calculate Cohen's $h$ \citep{cohen1988statistical} for comparing proportions and Cohen's kappa \citep{cohen1960coefficient} for measuring agreement beyond chance. We translate effect sizes into real-world impact by expressing differences as additional correct evaluations per 1000 simulations, providing direct insight into practical implications.

\subsubsection{Robustness Validation}
\label{sec:robustness}

We validate our findings through bootstrap confidence intervals (10,000 iterations) \citep{efron1979bootstrap} and permutation tests \citep{good2000permutation} to ensure results are not artifacts of specific methodological choices. Performance consistency is assessed through coefficient of variation analysis, and we verify convergence across multiple scoring system variants to confirm the stability of our conclusions.

\section{Simulation Study}
\label{sec:simulation}

\subsection{Context and Motivation}

The quality of voice AI testing platforms fundamentally depends on their ability to generate realistic simulations that accurately reflect production scenarios. While evaluation quality (see Section ~\ref{sec:evaluation}) determines whether platforms can correctly score conversations, simulation quality determines whether those conversations represent real life user interactions in the first place. Poor simulation fidelity undermines the testing process, independent of evaluation accuracy.

This study addresses a critical gap in voice AI testing: the systematic evaluation of how well testing platforms can simulate realistic customer interactions under controlled conditions. By comparing three leading platforms---Evalion, Coval, and Cekura---we establish empirical evidence for simulation quality differences that directly impact the reliability of automated testing in production environments.

\subsection{Study Objectives}

As outlined in Section ~\ref{subsec:simulation-quality-concept-approach}, evaluating simulation quality ultimately amounts to assessing the quality of the platform’s testing agents. This study seeks to quantify the relative performance of these testing agents by measuring how effectively they simulate realistic conversations under controlled conditions.

In the context of this study, a simulation is always built around two key inputs: a scenario and a persona. The scenario describes the conditions of the test, such as the objectives or constraints of the conversation from the testing agent’s perspective. The persona specifies the personality traits the testing agent must represent. Both scenario and persona are provided as human-written prompts that all three providers compared in the present study implement as inputs in their platforms.

When evaluating testing agents with this framework, we define quality according to two main aspects.
\begin{enumerate}
\item First, adherence to the assigned test conditions: this includes both the ability of the testing agent to follow the scenario parameters set for the test as well as the ability to accurately represent the testing persona it has been assigned.
\item Second, perceived human-likeness: this dimension captures how natural, coherent, and human-like the testing agent sounds during interactions.
\end{enumerate}

Through systematic evaluation of these dimensions, we can compare the relative simulation performance of the three tested providers and highlight their respective strengths and weaknesses against each other.

\subsection{Experimental Design}

\subsubsection{Evaluation Setup}

To ensure fair comparison across providers, we designed a controlled experimental framework where all three platforms execute identical test cases. Each platform received the same scenario descriptions and persona specifications, implemented as human-written prompts directly in their respective systems. All simulations for all providers were run against the same subject agent, an enterprise-ready inbound customer support agent provided by Sei Right \citep{sei}.

For each combination of scenario and persona, we conducted pairwise comparisons between all three providers, resulting in three comparison surveys per simulation. Human evaluators rated these comparisons across multiple metrics, with scores subsequently aggregated to produce final rankings. This tournament-style approach ensures that each provider's simulations are evaluated against every other provider under identical conditions.

\subsubsection{Methodological Approach: Pairwise Comparisons}

We choose to use direct comparisons between audio files rather than independent ratings in isolation. The methodology is inspired by established evaluation practices in machine learning research. Platforms such as LMArena \citep{lmarena} have popularized the use of pairwise comparisons to evaluate large language models, showing that head-to-head judgments yield clearer and more consistent rankings than scalar ratings. Similar findings have been observed in psychometrics and human-computer interaction research, where comparative judgments often lead to higher inter-rater agreement \citep{thurstone}.

By applying this framework to voice agent evaluation, we ensure that human judgments are robust, interpretable, and aligned with best practices in AI system benchmarking.

\subsection{Metrics and Measurement Framework}

\subsubsection{Primary Components}

The evaluation relies on three main components, each designed to capture a different yet complementary dimension of simulation quality:

\begin{itemize}
    \item \textbf{Scenario Adherence:} Assesses whether the testing agent’s behavior and dialogue remain consistent with the given scenario.\\
    \textit{Focus}: Accuracy of facts, relevance of requests, and coverage of scenario requirements.
    \item \textbf{Persona Adherence:} Evaluates how well the testing agent reflects the assigned persona throughout the interaction.\\
    \textit{Focus}: Emotional cues, tone of voice, word choice, and behavioral consistency.
    \item \textbf{Human Naturalness:} Measures how convincingly human the testing agent sounds, focusing on vocal and conversational qualities.\\
    \textit{Focus}: Prosody, intonation, rhythm, turn-taking, and overall naturalness.
\end{itemize}

These three dimensions were selected because they directly align with the fundamental requirements of simulation-based evaluation in conversational AI. \textbf{Scenario Adherence} and \textbf{Persona Adherence} correspond to the two main inputs that current providers allow users to define when constructing test cases, i.e., the conditions of the scenario and the behavioral profile of the persona. It is therefore essential to measure the degree to which testing agents remain faithful to these specifications. \textbf{Human Naturalness}, on the other hand, addresses a broader industry concern: simulations are most valuable when they approximate real human interactions. Evaluating how convincingly human the testing agents sound ensures that the tested systems are challenged under conditions that closely resemble authentic customer interactions.

Each of these dimensions is scored independently with a standardized range. See details in Section ~\ref{sec:scoring-methodology}.

\subsubsection{Composite Scoring}

To provide an overall assessment of simulation quality, we computed a weighted composite score:

\begin{equation}
\text{\it{Overall Score}}= 0.4 \cdot SA + 0.3 \cdot HN + 0.3 \cdot PA
\end{equation}

where \textit{SA} represents Scenario Adherence, \textit{HN} represents Human Naturalness, and \textit{PA} represents Persona Adherence.

This weighting scheme reflects industry priorities, particularly the needs of developers who form the core user base of these providers. For them, scenario fidelity is critical, as it ensures that simulations accurately reflect the test cases they define. For transparency, we additionally report the unweighted average of the overall score next to the weighted overall score.

\subsubsection{Metric Breakdown}
\label{subsec:sim-metric-breakdown}

To enable reliable human evaluation, each primary component was decomposed into specific, measurable metrics. Rather than asking vague questions like ``How human did the agent sound?`` or ``How well did the agent adhere to the scenario?``, evaluators assessed concrete aspects such as intonation, pauses, emotional tone, and response coherence.

This approach has three key benefits:

\begin{enumerate}
    \item \textbf{Reduce cognitive load} for human participants by asking them to evaluate more specific and independent aspects of the conversation.
    \item \textbf{Increase measurement consistency}. By focusing on more specific metrics, responses become less subjective and more comparable across different evaluators and simulations.
    \item \textbf{Construct validity}. By measuring a wider variety of metrics, it allows us to examine how different aspects of human speech and prosody relate to one another and how they together explain the overall quality of conversational voice AI agents.
\end{enumerate}

In the context of the conducted surveys, each metric was linked to a specific survey question. Human participants were asked to compare a pair of simulations of two providers and selected the best simulation according to each metric (allowing for ties too).

Below is the full metric overview, with their respective survey questions presented to each participant.

\textbf{Scenario Adherence} breakdown
\begin{itemize}
    \item Completeness: \textit{Which audio is more aligned with the following statement?} \\
    \textit{The customer talked about all the issues in the scenario.}
    \item Accuracy: \textit{Which audio is more aligned with the following statement?} \\
    \textit{The customer explained their problems clearly and correctly.}
    \item Goal pursuit: \textit{Which audio is more aligned with the following statement?} \\
    \textit{The customer talked about all the issues in the scenario.}
    \item Hallucinations: \textit{Which audio is more aligned with the following statement?} \\
    \textit{The customer brought up problems that were not in the scenario.}
    \begin{itemize}
        \renewcommand{\labelitemii}{$\circ$}
        \item Note that, in this case, being marked as the winner is actually a loss, since it means that the testing agent hallucinated. Hence, we exchange winners and losers in the data postprocessing to score this metric.
    \end{itemize}
    \item Overall Adherence: \textit{Overall, in which audio did the customer act more according to the scenario?}
\end{itemize}

The selected metrics provide a comprehensive coverage of what it means for a testing agent to remain consistent with the given scenario. \textbf{Completeness}, \textbf{accuracy}, and \textbf{goal pursuit} ensure that the testing agent fully addresses the scenario requirements without omission or misrepresentation. The \textbf{hallucinations} metric safeguards against deviation by penalizing the introduction of irrelevant issues. Finally, \textbf{overall adherence} provides a holistic check that integrates these aspects. Together, they reflect whether the agent’s dialogue is faithful to the scenario in terms of factual accuracy, relevance, and coverage.

\textbf{Human Naturalness} breakdown
\begin{itemize}
    \item Voice Naturalness: \textit{In which audio the customer’s voice sounds more human, not robotic?}
    \item Speaking Flow: \textit{In which audio the customer talks at a more natural pace – smooth, not too fast, slow, or broken up?}
    \item Tone and Emotion: \textit{In which audio the customer’s tone makes their feelings more clear (e.g., happy, annoyed, stressed, confused)?}
    \item Word Choice: \textit{In which audio the customer uses more simple, natural words that are easy to understand?}
    \item Response Fit: \textit{In which audio the customer’s replies make more sense and better follow the conversation?}
    \item Overall Naturalness: \textit{Overall, in which audio the customer sounds more like a real person?}
\end{itemize}

These metrics capture key elements that make a voice interaction sound convincingly human. \textbf{Voice naturalness} and \textbf{speaking flow} assess the prosody, intonation, and rhythm that distinguish human from robotic speech. \textbf{Tone and emotion} reflects the agent’s ability to convey affect, an essential part of natural communication. \textbf{Word choice} and \textbf{response fit} evaluate how the agent uses language and context to maintain coherent, meaningful dialogue. Finally, overall naturalness provides a holistic judgment that integrates these cues. Together, these metrics comprehensively assess the prosodic, emotional, and conversational qualities central to human-likeness.

\textbf{Persona Adherence} breakdown

Persona Adherence is measured using traits that together approximate a realistic characterization of a customer persona. Depending on the assigned persona, each rating is considered a win or a loss differently. For example, if the persona is defined as stressed and a simulation is rated as calmer, this counts as a loss, and vice versa. The traits include:

\begin{itemize}
    \item Emotional Tone: \textit{In which audio the customer sounds more calm?}
    \item Cooperation: \textit{In which audio the customer is more cooperative in solving the problem?}
    \item Communication Style: \textit{In which audio the customer explains themselves more clearly and concisely?}
    \item Respect: \textit{In which audio the customer uses more polite and respectful language?}
    \item Patience: \textit{In which audio the customer sounds more patient and willing to wait?}
\end{itemize}

\textbf{Emotional tone}, \textbf{cooperation}, and \textbf{communication style} address how the customer expresses themselves and engages in problem-solving. \textbf{Respect} and \textbf{patience} capture interpersonal and behavioral dimensions that shape the overall impression of the persona. Taken together, these traits provide a multi-faceted view of personality, allowing us to evaluate whether the testing agent consistently reflects the intended persona throughout the interaction.

Noteworthy, for persona adherence we do not include a dedicated question on overall adherence since that would require human evaluators to be presented with a persona prompt, which is rather complex and can add fatigue to the survey. Also, some personas are not rated on certain metrics, because the traits scored are not applicable. This is detailed in the \ref{subsec:test-personas} section below.

\subsection{Scoring Methodology}
\label{sec:scoring-methodology}

\subsubsection{Scoring systems}

Since survey participants only provided pairwise comparisons between simulations (i.e., choosing which one was better for each metric, or rating as a tie), a scoring system is needed to translate these individual comparison outcomes into an overall score for each metric and simulation. Without such a system, it would not be possible to aggregate the results into a fair and interpretable ranking.

To achieve this, we applied two of the most widely used and well-established scoring methods for aggregating pairwise comparisons \citep{pairwiseelo, pairwiseleague}:

\begin{itemize}
    \item \textbf{League system}: Each comparison is treated like a match. A “win” assigns the chosen simulation 1 point, while a “loss” assigns 0 points. This method is simple, transparent, and directly reflects the number of times a simulation was preferred.
    \item \textbf{Elo system}: Adapted from chess ratings, this method accounts not only for wins and losses but also for the relative strength of the competing simulations. This allows strong simulations to be rewarded more for beating other strong ones, while losses against weaker ones have a greater penalty \citep{elorating}.\\
    In our implementation, simulations start with a baseline Elo rating of 1500, and their rating is adjusted after each comparison using a K-factor of 32.
\end{itemize}

\subsubsection{Tie handling}

For each these two scoring systems, we also implemented two variants for handling ties between simulations.

\begin{itemize}
    \item \textbf{Including ties}: In the league system, a tie yields 0.5 points for each simulation. In the Elo system, ties are replaced by two artificial matches—one win for each system—in random order since Elo is not commutative \citep{hamilton2024elo}.
    \item \textbf{Excluding ties}: Ties are ignored entirely.
\end{itemize}

\subsubsection{Metric Aggregation}
In order to implement the aggregated Scenario Adherence, Persona Adherence and Human Naturalness scores, we also implemented two alternative aggregation approaches:

\begin{itemize}
    \item \textbf{PCA aggregation}: Using Principal Component Analysis (PCA), the first component is taken as the score. This ensures the score captures the greatest variance across the metrics \citep{pcaanalysis} in each dimension.
    \item \textbf{Hybrid approach}: PCA is applied only for Persona Adherence, while Scenario Adherence and Human Naturalness rely directly on their respective Overall Adherence and Overall Naturalness metrics. Note that, for Persona Adherence, there is no question about overall adherence and hence we must use an aggregate over its metrics.
\end{itemize}

\subsubsection{Normalization}

To enable direct comparison, combination, and aggregation of all metrics, we normalized them to a common scale. Specifically, we applied min–max normalization \citep{minmaxscaling} across providers and simulations, rescaling each metric to the range 0–100 for consistency and simplicity.

\begin{equation}
X_{\text{Scaled}} = 100 \cdot \frac{x - \min(x)}{\max(x) - \min(x)}
\end{equation}

\subsubsection{Post analysis and metric validation}

Combining the scoring systems (League vs. Elo), tie handling (include vs. exclude), and aggregation methods yields a total of eight scoring variants. These different variants were all analyzed, including resulting ratings on the primary components and the subsequent overall scores.

As part of the post-analysis in the Results section, we validate the \textbf{Overall Scenario Adherence} and \textbf{Overall Human Naturalness} metrics by performing linear regressions against their respective complementary metrics inside Scenario Adherence and Human Naturalness. This step verifies that these overall measures can be reliably explained by the underlying dimensions, supporting their use as single-point representations of Scenario Adherence and Human Naturalness respectively.

In addition, we examine the correlation between the different scoring variants (League vs. Elo, inclusion or exclusion of ties, and aggregation methods). This analysis allows us to assess whether methodological choices substantially impact the final outcomes. While we use these correlations to confirm the consistency of our results, we also report the scores and provider rankings for all scoring variants to ensure completeness and transparency.

\subsection{Experimental Setup}
\label{sec:simulation-experimental-setup}

As discussed previously, each simulation is an audio file linked to a given scenario and persona definition. A simulation is generated for each provider with the same scenario and persona prompts, calling the same subject agent. Next, each simulation is rated against the other two providers across the metrics presented in Section ~\ref{subsec:sim-metric-breakdown}.

In total, this combination was run across $T_s = 15$ scenarios and $P_s = 3$ personas. Hence, a total of $S_s = 45$ simulations for each provider were rated against each other.

Noteworthy, we selected 15 scenarios to provide sufficient coverage across a varying degree of difficulty while keeping the study tractable in terms of simulation and survey costs. The inclusion of 3 distinct personas ensured variation in customer communication styles and behavioral traits, allowing us to test whether providers could consistently adapt to different personalities. See the subsections ~\ref{subsec:test-scenarios} and ~\ref{subsec:test-personas} below for details on how these are built.

\subsubsection{Participants}
Each simulation pair was evaluated by $N_s = 10$ participants recruited through Prolific \citep{prolific2025}. To assess whether a larger participant pool was necessary, we conducted an additional analysis: from each survey we drew a random subset of 5 participants and recomputed the ranked scores. The resulting rankings were highly consistent with those obtained using all 10 participants, indicating that increasing the number of participants would not meaningfully alter the outcomes (see Section ~\ref{subsubsec:sim-number-participants-discusson}). On this basis, we concluded that surveying 10 participants per simulation pair was sufficient, while further scaling would only increase study costs without providing additional benefit. For context, a full survey with the $S_s = 45$ simulations, $C = 3$ providers and $N_s = 10$ participants, has an approximate cost of USD 2,500.

In total, each participant answered $M_s = 16$ questions per survey (see Section ~\ref{subsec:sim-metric-breakdown}).

The total number of surveys amounts to ${C \choose 2} \cdot T_s \cdot P_s = 135$. This leads to a total of ${C \choose 2} \cdot N_s \cdot M_s \cdot T_s \cdot P_s = 21600$ data responses for the study (see Section ~\ref{subsec:pairwise-comparison-structure} for details).

\subsubsection{Test Scenarios}
\label{subsec:test-scenarios}

A key challenge in designing this study was to ensure that scenarios were sufficiently complex to expose potential failures in the simulations, such as incomplete coverage, factual inaccuracies, or hallucinated content by the testing agent. To achieve this, we used a large language model (LLM) to generate 30 candidate scenarios based on the tested agent prompt, including paths outside the prompt definition of the subject agent. This approach ensured that the agent was evaluated on scenarios that probe its capacity to handle novel, more challenging scenarios.

Additionally, we wanted each scenario to carry an explicit indication of its difficulty. That is, how hard it is to score well on Scenario Adherence, or in other words, how hard it is to simulate successfully. This difficulty flag would allow us not just to ensure we have complex enough scenarios but also analyze how providers perform across the spectrum of scenario difficulty.

To select the final subset of scenarios and assign difficulties, we ran 3 simulations per provider and scenario using a calm and cooperative persona. This led to 270 simulations in total (3 providers with 3 simulations each on 30 candidate scenarios). The transcripts of these simulations were then evaluated by an LLM-as-a-judge prompt, which scored the Scenario Adherence metrics on a Likert scale \citep{likert1932technique}.

Using these scores and their variance across the whole set of simulations per scenario, we manually selected a balanced subset of $T_s = 15$ scenarios: 4 classified as easy, 6 as medium, and 5 as hard. The final subset is included in the \textbf{Appendix~\ref{app:test-sceanrios}}. For reference, follow two examples of scenarios, one of easy difficulty and one hard:

\begin{itemize}
\item \textbf{Easy:} \textit{You already activated your account but forgot your password. You want help using the `Forgot Password' flow, using the temporary password email, and then creating a new password.}
\item \textbf{Hard:} \textit{You are a settlement agent. You shared a closing with a notary. The notary claims they completed their profile but you still can't see their phone number in the Contacts tab. You also need to know if you can set a default notary for all your closings.}
\end{itemize}

\subsubsection{Tested Personas}
\label{subsec:test-personas}

We chose to test $P_s = 3$ different customer personalities to assess how well the systems could adapt to distinct communication styles, patience, politeness\ldots{} Each persona emphasizes different conversational challenges and expectations. Brief descriptions are provided below, with the full tested prompts included in \textbf{Appendix~\ref{app:personas-prompt}}:

\begin{itemize}
\item \textbf{Olivia}: Blunt, demanding, and easily frustrated communicator who expresses anger quickly and pushes aggressively for immediate resolution and validation.
\item \textbf{Sophia}: Short on time, direct, and efficient communicator who values quick resolution and clear, actionable information.
\item \textbf{Chloe}: Calm, polite, and cooperative communicator who seeks fair, clear solutions through understanding and constructive dialogue.
\end{itemize}

These three personas were chosen to reflect a diverse yet realistic range of customer types that a support agent is likely to encounter in practice. \textbf{Olivia} represents high-frustration and high-demand interactions, where emotional intensity and pressure for rapid resolution test the robustness of a system. \textbf{Sophia} embodies efficiency-driven customers, emphasizing clarity, brevity, and actionable responses under time constraints. \textbf{Chloe}, by contrast, reflects cooperative and patient customers, where the system must sustain a polite and constructive dialogue. Together, these personas span a spectrum of emotional tone, communication style, and interactional demands, providing sufficient variability to stress-test simulation systems under conditions that approximate real-world customer diversity in support contexts.

Regarding how the persona affects the scoring of each metric, we applied the following criteria based on the definition of the personality and each metric:

\begin{itemize}
\item For Emotional Tone, a better Olivia representation is expected to lose on this metric, and same for Sophia. On the other hand, a better representation of Chloe is expected to win.
\item For Cooperation, Olivia is expected to lose whereas Chloe and Sophia are expected to win.
\item For Communication Style, Sophia is expected to win whereas Chloe is expected to lose. Olivia on the other hand is not ranked on this metric since we consider it irrelevant for her evaluation.
\item For Respect, Chloe is expected to win while Olivia is expected to lose. Sophia on the other hand is not ranked on this metric since we consider it irrelevant for her evaluation.
\item For Patience, Olivia and Sophia are expected to lose whereas Chloe is expected to win.
\end{itemize}

This criteria is applied in the data postprocessing by exchanging winners and losers accordingly, similarly to what we do for the Hallucinations metric in Scenario Adherence.

\subsubsection{Survey tooling, anonymisation, and randomisation}

To run the study, we built a custom audio survey tool, ensuring all the methodological requirements for fair and consistent evaluation were met. Details on the specifications of this tool are presented in Section ~\ref{subse:survey-infra}.

For the simulation study, we configured the tool to present pairwise \textit{comparison} surveys. Each survey included the scenario description, the persona prompt, and the two corresponding audio simulations to be judged. To preserve anonymity and reduce bias, the provider identities (Evalion, Coval, Cekura) were never disclosed to participants. Furthermore, the placement of each provider's simulation was randomized: audio samples could appear on either the left or right side of the comparison interface. This ensured that participants' judgments reflected only the quality of the simulations, not preconceived expectations about the providers.

A screenshot of a survey using this tool is included in Figure~\ref{fig:survey-tool} (a). For details on the recruiting of human participants for the survey, as well as their eligibility criteria, refer to section \ref{subsubsec:crowsourced-humans-prolific}.

\subsection{Simulation Study Results}

In this section, we present the results of our evaluation based on the framework introduced in the Section ~\ref{sec:scoring-methodology} section. We computed the three primary scores, \textbf{Scenario Adherence}, \textbf{Persona Adherence}, and \textbf{Human Naturalness}, using eight alternative scoring variants derived from the surveyed metrics described in previous sections. We also report an \textbf{Overall Score}, defined as a weighted average of the three scores (40\% Scenario Adherence, 30\% Persona Adherence, 30\% Human Naturalness), and include an unweighted average (same weight for all 3 scores) version of it for completeness.

For each provider, we report the \textbf{average scores for all three primary components and for the overall score}, aggregated across the \textbf{45 test cases} (15 scenarios × 3 personas), with each pairwise comparison survey evaluated by \textbf{10 human participants}, yielding to a \textbf{total of 21,600 human judgments} aggregated in this analysis.

For simplicity, the main text includes results from a single scoring variant. However, we confirmed that rankings and relative performances remain consistent across all eight scoring variants. The aggregate scores for all eight variants are included in the \textbf{Appendix~\ref{app:socring-variants-res}} for completeness. The respective scores, for all metrics and primary components, on each scenario, persona and provider, are also referenced in \textbf{Appendix~\ref{app:socring-variants-res}}. 

In addition, we also report results disaggregated by scenario difficulty level (easy, medium, hard) to examine whether provider performance is consistent across different levels of conversational complexity.

The section also includes a brief discussion on the number of participants, and the adequacy of using ten participants per survey in the presented study.

Finally, it presents a post-analysis validating the overall scenario adherence and overall human naturalness metrics through regression on their complementary measures, and examining correlations across scoring variants to confirm that methodological choices do not substantially affect the outcomes.

All assets required to reproduce the study, including code, survey files, and human responses data, are referenced in the \textbf{Appendix~\ref{app:code-and-files}} section.

\subsubsection{Results}\label{sec:simulation-results}

Follows a summary table showing the aggregate scores for each provider under the league scoring system, including ties and using the overall adherence metric for scoring Scenario Adherence and Human Naturalness, combined with the PCA first component for the Persona Adherence. All other scoring variants used, with some small margin of variation, show similar rankings on aggregate (see Appendix \textbf{Appendix~\ref{app:socring-variants-res}} section).

\begin{table}[h!]
\centering
\begin{tabular}{lcccccc}
\hline
\textbf{Provider} & \textbf{Wins} & \textbf{Scenario} & \textbf{Human} & \textbf{Persona} & \textbf{Overall} & \textbf{Overall} \\
& \textbf{(of 45)} & \textbf{Adherence} & \textbf{Naturalness} & \textbf{Adherence} & \textbf{Score} & \textbf{Score (Unweighted)} \\
\hline
Evalion & \textbf{26} (58\%) & \textbf{63.72} & \textbf{62.11} & \textbf{57.29} & \textbf{61.31} & \textbf{61.04} \\
Coval & 12 (27\%) & 49.11 & 46.78 & 50.82 & 48.92 & 48.90 \\
Cekura & 7 (15\%) & 37.17 & 41.11 & 52.79 & 43.04 & 43.69 \\
\hline
\end{tabular}
\caption{Overall simulation performance across providers}
\label{tab:sim_overview}
\end{table}

Table~\ref{tab:sim_overview} includes the average scores for each provider across the 45 combinations of scenario and persona. It also counts how many combinations each provider is winning based on the (weighted) Overall Score metric.

In aggregating the scores by provider and scenario difficulty level, we observe a consistent ranking across easy, medium, and hard scenarios. Because the scores are based on min–max normalization, providers that rank similarly against one another at each difficulty level will also show similar final scores overall. It is important to note, however, that this does not imply the absolute quality of the simulations is the same across difficulty levels, only that their relative standing remains stable.

\begin{table}[h!]
\centering
\begin{tabular}{llcccccc}
\hline
\textbf{Provider} & \textbf{Difficulty} & \textbf{Wins} & \textbf{Scenario} & \textbf{Human} & \textbf{Persona} & \textbf{Overall} & \textbf{Overall} \\
& & & \textbf{Adh.} & \textbf{Nat.} & \textbf{Adh.} & \textbf{Score} & \textbf{Score (Unweighted)} \\
\hline
\multirow{3}{*}{Evalion} 
& Easy & 7 & 70.21 & 64.17 & 55.02 & 63.84 & 63.13 \\
& Medium & 11 & 58.33 & 60.28 & 58.89 & 59.08 & 59.17 \\
& Hard & 8 & 65.00 & 62.67 & 57.20 & 61.96 & 61.62 \\
\hline
\multirow{3}{*}{Coval} 
& Easy & 3 & 45.21 & 47.71 & 53.73 & 48.52 & 48.88 \\
& Medium & 4 & 49.44 & 46.25 & 48.67 & 48.25 & 48.12 \\
& Hard & 5 & 51.83 & 46.67 & 51.06 & 50.05 & 49.85 \\
\hline
\multirow{3}{*}{Cekura} 
& Easy & 2 & 34.58 & 38.13 & 52.15 & 40.91 & 41.62 \\
& Medium & 3 & 42.22 & 43.47 & 53.34 & 45.93 & 46.34 \\
& Hard & 2 & 33.17 & 40.67 & 52.64 & 41.26 & 42.16 \\
\hline
\end{tabular}
\caption{Performance by scenario difficulty level}
\label{tab:sim_difficulty}
\end{table}

\begin{figure}[h!]
\centering
\includegraphics[width=0.7\columnwidth]{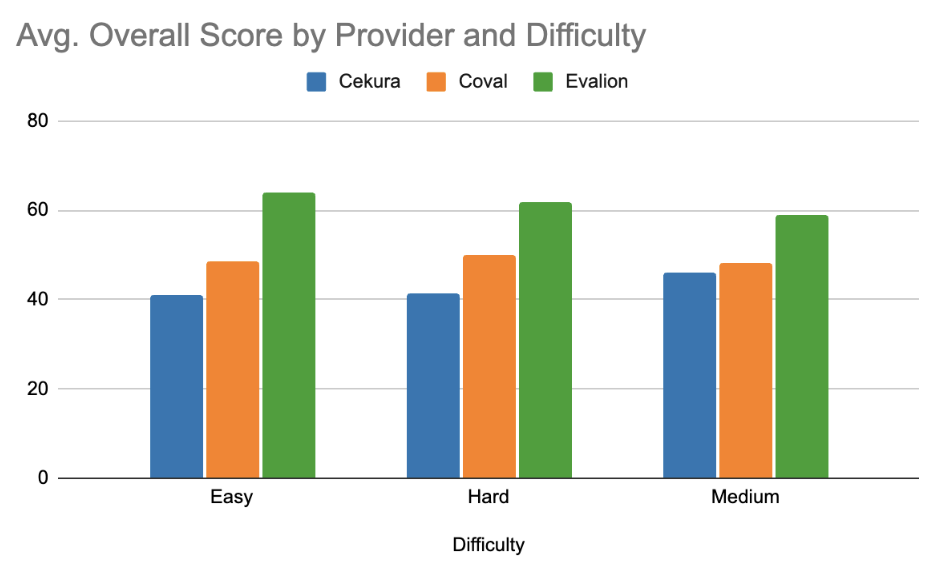}
\caption{Simulation (weighted) Overall Scores by provider and difficulty.}
\label{fig:sim-results-by-difficulty}
\end{figure}

The aggregated overall score for Table~\ref{tab:sim_difficulty} is also represented in Figure~\ref{fig:sim-results-by-difficulty}.

\subsubsection{Discussion on the number of human participants}
\label{subsubsec:sim-number-participants-discusson}

An additional analysis was conducted to evaluate whether the number of participants per survey was sufficient to provide reliable results. To test this, we repeated the scoring procedure using only a random subset of \textbf{5 participants} from each survey, and recalculated the average overall scores across metrics and providers. For this exercise, we just used the league scoring including ties and the overall adherence metric rating for Scenario Adherence and Human Naturalness. The results for this exercise are reported in Table~\ref{tab:sim_overview_5p}.

The resulting aggregates were highly similar to those obtained with the full set of 10 participants, supporting the conclusion that increasing the participant pool beyond 10 would not significantly change the findings. This validates our choice of 10 participants per pair as both cost-effective and methodologically sound.

\begin{table}[h!]
\centering
\begin{tabular}{lccccc}
\hline
\textbf{Provider} & \textbf{Scenario} & \textbf{Human} & \textbf{Persona} & \textbf{Overall} & \textbf{Overall} \\
 & \textbf{Adherence} & \textbf{Naturalness} & \textbf{Adherence} & \textbf{Score} & \textbf{Score (Unweighted)} \\
\hline
Evalion & 64.78 & 63.33 & 57.54 & 62.17 & 61.88 \\
Coval & 47.22 & 46.22 & 50.50 & 47.91 & 47.98 \\
Cekura & 38.00 & 40.44 & 51.92 & 42.91 & 43.45 \\
\hline
\end{tabular}
\caption{Overall simulation performance across providers (with 5 participants)}
\label{tab:sim_overview_5p}
\end{table}

\subsubsection{Post Analysis and Metrics Validation}

In order to validate the metrics collected and to assess the impact of the scoring variant selection, this section reports a correlation analysis conducted across all eight scoring variants, as defined by the combination of scoring systems, tie-handling rules, and aggregation methods. We further evaluate the validity of the \textbf{Overall Adherence} and \textbf{Overall Naturalness} metrics by regressing them against their complementary metrics inside Scenario Adherence and Human Naturalness respectively. Both analysis provide assurance that the aggregated metrics are robust representations of the underlying dimensions and that the choice of scoring variant does not materially affect final rankings.

\textbf{Regression Analysis}

For the \textbf{Scenario Adherence}, the regression analysis demonstrated that the \textbf{Overall Adherence} score can be largely explained by the other four metrics, with an \textbf{\textit{R²} = 0.893}. This suggests that participants consistently inferred overall adherence from the rest of surveyed questions.

Similarly, for \textbf{Human Naturalness}, a regression model predicting the \textbf{Overall Naturalness} score from the other metrics achieved an \textbf{\textit{R²} = 0.873}, indicating again that participants also inferred their overall human naturalness rating from other surveyed metrics.
Both regression analyses from above are detailed in Appendix~\ref{app:simulation_regression}.

\textbf{Scoring Variants Correlation Analysis}

\begin{figure}[h!]
\centering
\includegraphics[width=0.7\columnwidth]{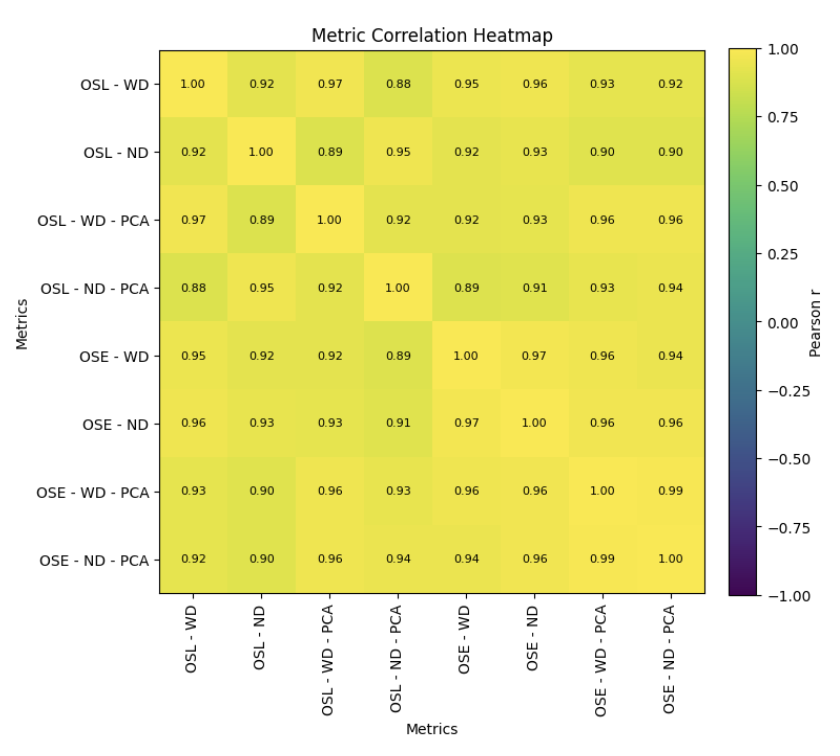}
\caption{Cross Correlations heatmap across scoring variants.}
\label{fig:correlation-scoring-variants}
\end{figure}

Finally, we compared the results obtained under the eight scoring variants (League vs. Elo, with/without ties, PCA vs. hybrid aggregation). For that, we computed the cross correlation for the overall score between all the scoring variants, reported in Figure~\ref{fig:correlation-scoring-variants}.

The correlation analysis revealed \textbf{high consistency across all scoring methods}, showing that provider rankings are robust to the choice of scoring system. This suggests that the final results are not sensitive to methodological differences in score computation, with the scoring variants analysed in this study.

Scoring Variant abbreviations in Figure~\ref{fig:correlation-scoring-variants}:
\begin{itemize}
    \item OSL - WD: Overall Score with League Scoring including ties.
    \item OSL - ND: Overall Score with League Scoring excluding ties.
    \item OSL - WD - PCA: Overall Score with League Scoring including ties and scoring Scenario Adherence and Human Naturalness with PCA first component.
    \item OSL - ND - PCA: Overall Score with League Scoring excluding ties and scoring Scenario Adherence and Human Naturalness with PCA first component.
    \item OSE - WD: Overall Score with Elo Scoring including ties.
    \item OSE - ND: Overall Score with Elo Scoring excluding ties.
    \item OSE - WD - PCA: Overall Score with Elo Scoring including ties and scoring Scenario Adherence and Human Naturalness with PCA first component.
    \item OSE - ND - PCA: Overall Score with Elo Scoring excluding ties and scoring Scenario Adherence and Human Naturalness with PCA first component.
\end{itemize}

\subsubsection{Summary of Simulation Results}

The simulation study establishes clear performance hierarchy among platforms. Evalion demonstrates superior simulation quality across all 3 primary components, with particular strength in \textbf{Scenario Adherence} and \textbf{Human Naturalness}. The 61.31 overall score represents 24.7\% improvement over Coval and 41.7\% over Cekura. This advantage is consistent across scenario difficulty levels.

These findings provide practitioners with empirical guidance for platform selection based on their specific testing requirements and scenario complexity.
\section{Evaluation Study}
\label{sec:evaluation}

\subsection{Context and Motivation}

The automation of voice AI agent testing represents a critical challenge in scaling quality assurance for conversational systems. While the simulation study demonstrated platform differences in generating realistic test conversations, the ability to accurately \emph{evaluate} those conversations determines whether automated testing can reliably replace or augment human assessment in production environments.

This evaluation study directly addresses the fundamental question underlying automated testing platforms: Can automated evaluation systems accurately replicate human assessment of conversational quality? The answer determines whether organizations can confidently deploy automated testing in production environments, potentially evaluating thousands of conversations daily rather than the dozens or hundreds feasible with human reviewers (not to mention the much higher cost associated with human reviewers). By establishing empirical evidence for the gap between automated and human evaluation, this study provides the quantitative foundation necessary for informed adoption of automated testing platforms.

The primary objective of this evaluation study is to quantitatively assess the accuracy and reliability of automated voice AI evaluation platforms by comparing their model judges (LLM-as-a-judge) against human-established ground truth. Building upon the simulation quality findings, we now examine whether platforms can accurately assess conversation quality once simulations are generated. Specifically, we aimed to:

\begin{enumerate}
    \item Compare model judge performance across platforms by evaluating how different testing platforms' automated evaluation systems perform when analyzing identical conversation transcripts, focusing on content and meaning-related metrics.
    \item Establish human-validated ground truth through systematic human evaluation of voice AI agent interactions, capturing both binary and continuous performance metrics.
    \item Quantify the evaluation gap by measuring the performance difference between three commercial evaluation providers (Evalion, Cekura, and Coval) against human judgment using standardized metrics.
    \item Identify platform strengths and weaknesses to determine which metrics and platforms demonstrate highest concordance with human assessment, informing best practices for automated testing.
\end{enumerate}

The study leverages the observability API feature available in modern testing platforms, enabling fair comparison by ensuring all platforms evaluate identical conversation transcripts, thus eliminating variability from different simulation conditions.

\subsection{Metrics Used and Motivations}

\subsubsection{Metric Selection Criteria}
\label{subsec:eval-metrics-selected}

To ensure validity and practical relevance, we selected metrics based on four key criteria. First, metrics needed to be available as native features in at least one testing platform, ensuring we evaluated capabilities the industry has already identified as valuable. Second, metrics required relevance to real-world customer service evaluation, aligning with standard contact center KPIs such as CSAT (Customer Satisfaction Score), which maps to widely-used industry metrics, and Expected Outcome, which corresponds to first call resolution rates. Third, metrics needed measurability through both human and automated assessment to enable valid comparison between human and machine evaluation. Finally, we included both binary (pass/fail) and continuous (scaled) evaluation dimensions to capture the multi-faceted nature of conversation quality, where binary metrics capture critical success criteria and continuous metrics acknowledge the gradient nature of user experience.

The $M_e = 6$ metrics selected are: 
\begin{itemize}
    \item CSAT (Customer Satisfaction Score) is widely used in industry \citep{csat}.
    \item Expected Outcome: maps to first call resolution (FCR), a key contact center metric -- essentially it defines for a given scenario what is the expected behavior and outcome of the subject agent.
    \item Response Consistency: The subject agent is consistent in their answers and references. This addresses service reliability.
    \item Conversation Progression: The subject agent is trying to resolve the problems and move the conversation forward. This ensures efficient problem resolution.
    \item Appropriate Call Closure: The subject agent appropriately ends the call. This follows professional service standards.
    \item Avoid Repetition: The subject agent does not repeat itself many times. This prevents frustrating conversation loops.
\end{itemize}

\subsubsection{Platform Implementation}

Each platform offers different native implementations of evaluation metrics, with some requiring custom configuration. Table~\ref{tab:platform_metrics} shows the distribution of native versus custom implementations across platforms:

\begin{table}[h!]
\centering
\begin{tabular}{lccc}
\hline
\textbf{Metric} & \textbf{Coval} & \textbf{Cekura} & \textbf{Evalion} \\
\hline
CSAT & Custom & Native & Native \\
Appropriate Call Closure & Native & Custom & Custom \\
Avoid Repetition & Native & Custom & Custom \\
Conversation Progression & Native & Custom & Custom \\
Response Consistency & Custom & Native & Custom \\
Expected Outcome & Custom & Custom & Custom \\
\hline
\end{tabular}
\caption{Distribution of native vs custom metric implementation across platforms}
\label{tab:platform_metrics}
\end{table}

When native implementations were available, we prioritized their use to leverage each platform's optimized evaluation capabilities. For metrics requiring custom implementation, we developed standardized prompts (see Appendix~\ref{app:eval-prompts}) ensuring consistency across platforms.

The Expected Outcome metric required special handling due to the nature of observability testing. Since the Observability API evaluates arbitrary transcripts without scenario context, we embedded the corresponding scenario's expected outcome directly within the custom metric prompt for each transcript, ensuring all platforms received identical contextual information. Note that, even though all platforms offer native support for expected outcome for a simulation ran by the platform itself, we cannot use that functionality since we need to have the same transcript evaluated by all platforms.

\subsection{Experimental Design}
\label{subsec:eval-experimental-design}

\subsubsection{Golden Set Methodology}

The golden set methodology establishes human judgment as the criterion standard for evaluating automated systems. This approach provides an essential ground truth that otherwise a purely automated assessment cannot guarantee. Multiple human evaluators assessing the same conversations create a reliable benchmark reflecting real-world quality perception. The methodology enables systematic bias detection by comparing automated evaluations against human consensus, revealing consistent patterns where automated systems over- or under-estimate performance. Furthermore, it ensures fair cross-platform comparison through a common human-validated reference set, eliminating advantages from different internal benchmarks or evaluation philosophies. Ultimately, the gap between automated and human evaluation directly indicates whether automated systems can reliably replace or augment human quality assurance in production environments.

\subsubsection{Human Participants}

We recruited participants through Prolific \citep{prolific2025} with strict eligibility criteria to ensure evaluator quality. Our sample consisted of $N_e = 10$ participants per simulation. To prevent fatigue effects and ensure response diversity, we limited each participant to evaluating a maximum of three audio recordings.

For details on the recruiting of human participants for the survey, as well as their eligibility criteria, refer to section \ref{subsubsec:crowsourced-humans-prolific}. Note that this participant sample is fully independent from the participant sample of the simulation study in Section~\ref{sec:simulation}.  

\subsubsection{Conversation Corpus}

We selected $S_e = 60$ call simulations from the simulation study, choosing the highest-performing simulations based on Scenario Adherence scores to ensure high-quality baseline conversations with diverse scenario and persona combinations for evaluation. This yields a total of $N_e \cdot S_e = 600$ total evaluations.

\subsubsection{Data Collection Process}

Our evaluation protocol required participants to listen to at least 80\% of the audio recording in question. This was enforced by our custom survey tool which forced a minimum 80\% audio completion before enabling question responses. Participants answered six questions per simulation (one per metric) presented in randomized order to prevent order effects.

\subsubsection{Golden Set Construction and Consensus Analysis}

For each simulation ($S_e = 60$) and metric ($M_e = 6$) pair, we collected 10 human evaluations to establish ground truth. In terms of total data points, this yields a total of \textbf{3600 data points} collected to build the golden set for this study.

With this data collected, we defined consensus level as the percentage of evaluators agreeing on the majority answer. For instance, if 8 evaluators responded ``yes'' and 2 responded ``no,'' the consensus level would be 80\%. We considered consensus weak when fewer than 80\% of evaluators agreed on a metric for a specific simulation. Consensus analysis took into account only the binary metrics (5 in total), i.e., excluding CSAT. The rationale is that continuous metrics are more prone to subjectivity and hence we can contemplate some degree of variability.

Analysis of consensus patterns across all 60 recordings revealed varying levels of agreement. Table~\ref{tab:consensus_distribution} shows the distribution of recordings by their weak consensus count:

\begin{table}[h!]
\centering
\begin{tabular}{lcc}
\hline
\textbf{Weak Consensus Count} & \textbf{Number of Recordings} & \textbf{Percentage} \\
\hline
0 metrics & 45 & 75\% \\
1 metric & 10 & 16.67\% \\
2 metrics & 3 & 5\% \\
3 metrics & 2 & 3.33\% \\
\hline
\end{tabular}
\caption{Distribution of recordings by weak consensus metric count}
\label{tab:consensus_distribution}
\end{table}

75\% of the recordings achieved perfect consensus across all metrics, while an additional 16.67\% showed disagreement on only one metric. 

\textbf{Note} If we define consensus to be the average of agreeing votes across all metrics for a given simulation, then the consensus is 98.3\% (\url{https://docs.google.com/spreadsheets/d/15M52A6XXWfnE7JeM-l7HkggTc0s-mfebxxsfQgEvnks}).

In what follows, we retained all 60 simulations from the evaluation dataset rather than filtering to only the ones exhibiting high-consensus in human responses. This decision preserves the full diversity of conversational scenarios, reflects real-world conditions where some evaluations are inherently more ambiguous, provides a more robust test of platform capabilities across both clear and ambiguous cases, and maintains a larger sample size for stronger statistical power (60 simulations $\times$ 6 metrics = 360 observations per platform).

For transparency and to enable alternative analyses, Appendix~\ref{app:filtered-res} provides complete results using a filtered dataset of 45 recordings (excluding those with weak consensus on 1 or more metrics). The filtered analysis shows similar platform rankings but with slightly higher absolute accuracy scores, confirming that our inclusion of ambiguous cases provides a more conservative and realistic assessment of platform capabilities.

\subsection{Evaluation Results}\label{sec:evaluation-results}

\subsubsection{Ground Truth Validation}

To validate the reliability of human evaluations as ground truth, we analyzed inter-rater agreement across all metrics. The subject agent demonstrated strong overall performance, with human evaluators predominantly assigning positive ratings across most metrics, providing a meaningful baseline for platform comparison.

\begin{figure}[h!]
\centering
\includegraphics[width=0.5\columnwidth]{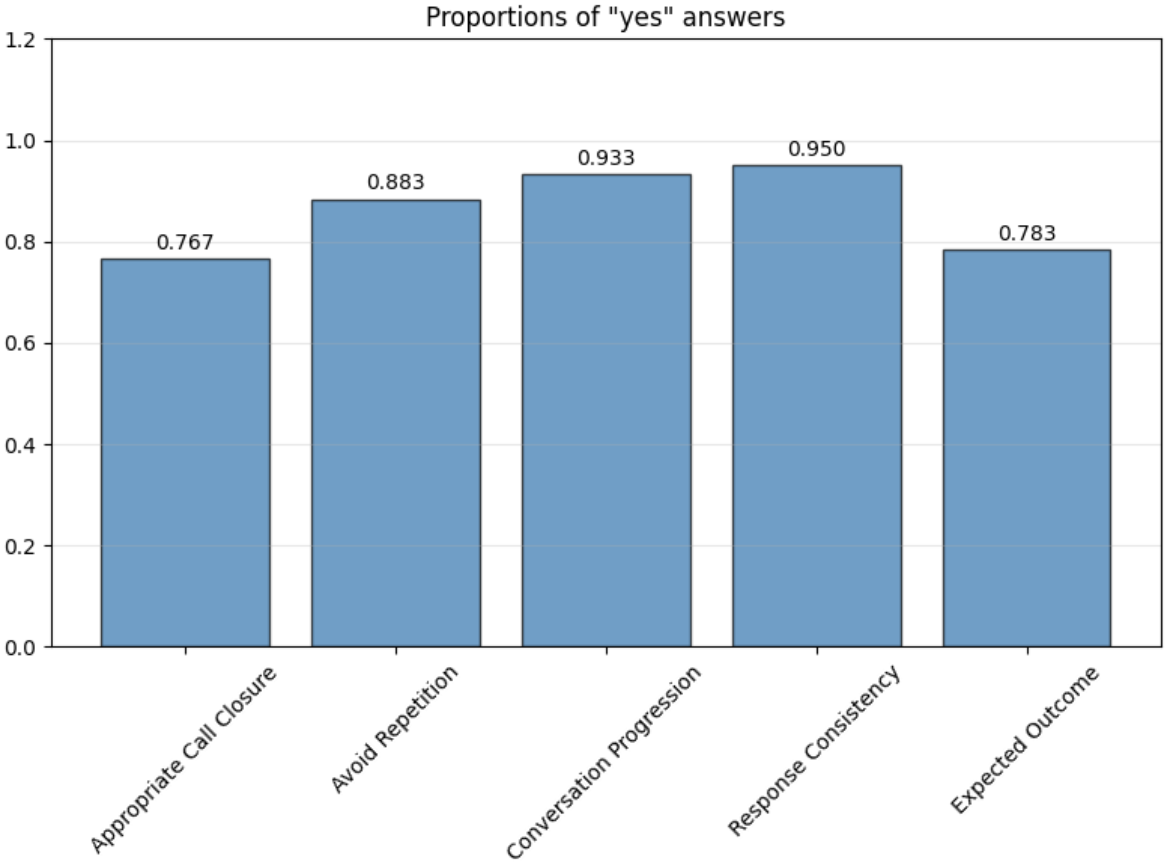}
\caption{Proportion of positive evaluations for binary metrics across all human evaluations. Values above each bar indicate the proportion of ``Yes'' responses (representing a positive score on the performance of the subject agent).}
\label{fig:binary_metrics_human}
\end{figure}

Figure~\ref{fig:binary_metrics_human} presents the proportion of positive responses for each binary metric across all 600 human evaluations (60 simulations $\times$ 10 evaluators). The binary metrics demonstrated both strong agent performance and evaluator consensus, with positive rates ranging from 76.7\% (Expected Outcome) to 95.0\% (Avoid Repetition), highlighting the maturity of the chosen subject agent, provided by Sei ~\citep{sei} This variation indicates evaluators discriminated between different quality dimensions rather than uniformly rating all aspects positively.

\begin{figure}[h!]
\centering
\includegraphics[width=0.5\columnwidth]{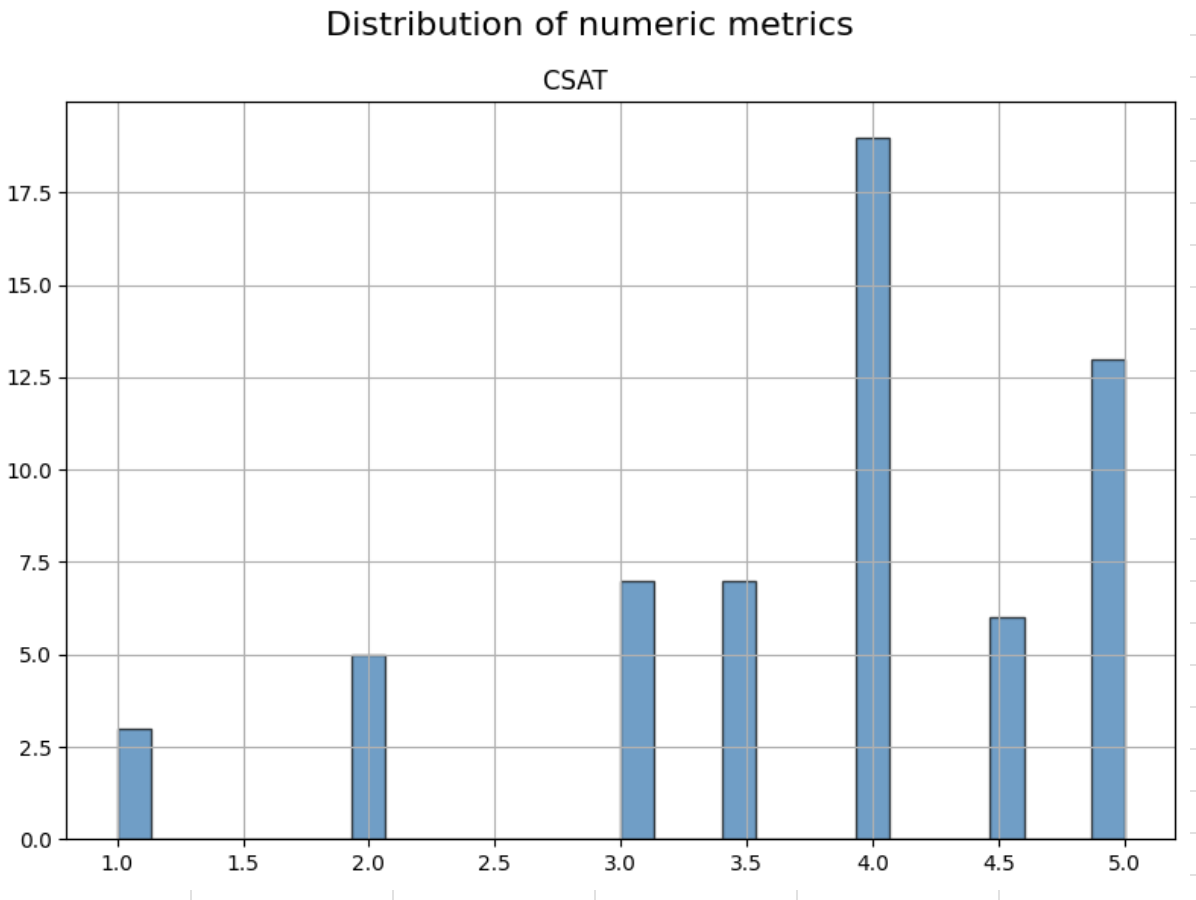}
\caption{Distribution of Customer Satisfaction (CSAT) scores across all 600 human evaluations.}
\label{fig:csat_distribution}
\end{figure}

The Customer Satisfaction (CSAT) distribution (Figure~\ref{fig:csat_distribution}) shows a strong positive skew with modal response at 4 and substantial frequency at 5, confirming that human evaluators perceived the conversations as satisfactory to highly satisfactory. The presence of lower scores in smaller frequencies demonstrates evaluator discrimination and validates that the assessment captured genuine quality variations rather than uniformly positive bias.

\subsubsection{Binary Metrics Performance}
\label{subsec:eval-bin-metrics-performance}

To evaluate how accurately each platform's evaluator aligns with human consensus, we computed standard classification metrics for each binary evaluation. For every platform-metric-simulation combination, we compared the platform's binary prediction against the human ground truth (established through majority consensus) to determine correctness. 

We employ four fundamental classification metrics \cite{sokolova2009systematic}. \textbf{Precision} measures the proportion of positive predictions that were correct: $P = TP/(TP + FP)$, where $TP$ denotes true positives and $FP$ denotes false positives. \textbf{Recall} (sensitivity) measures the proportion of actual positive cases correctly identified: $R = TP/(TP + FN)$, where $FN$ denotes false negatives. The \textbf{F1-score} provides the harmonic mean of precision and recall: $F_1 = 2PR/(P + R)$, balancing both metrics to penalize extreme values~\cite{powers2011evaluation}. \textbf{Accuracy} measures overall correct predictions: $A = (TP + TN)/(TP + TN + FP + FN)$, where $TN$ denotes true negatives \cite{fawcett2006introduction}.

In the context of the golden set curated in this study, with 76.7\%--95.0\% of human evaluations being positive across metrics (see Figure~\ref{fig:binary_metrics_human}), accuracy becomes misleading. For example, on Expected Outcome where 78.3\% of human evaluations were positive, a platform could achieve 78.3\% accuracy by simply predicting ``yes'' for every case. F1-score instead provides a balanced measure that penalizes both false positives and false negatives.

\begin{figure}[h!]
\centering
\includegraphics[width=\columnwidth]{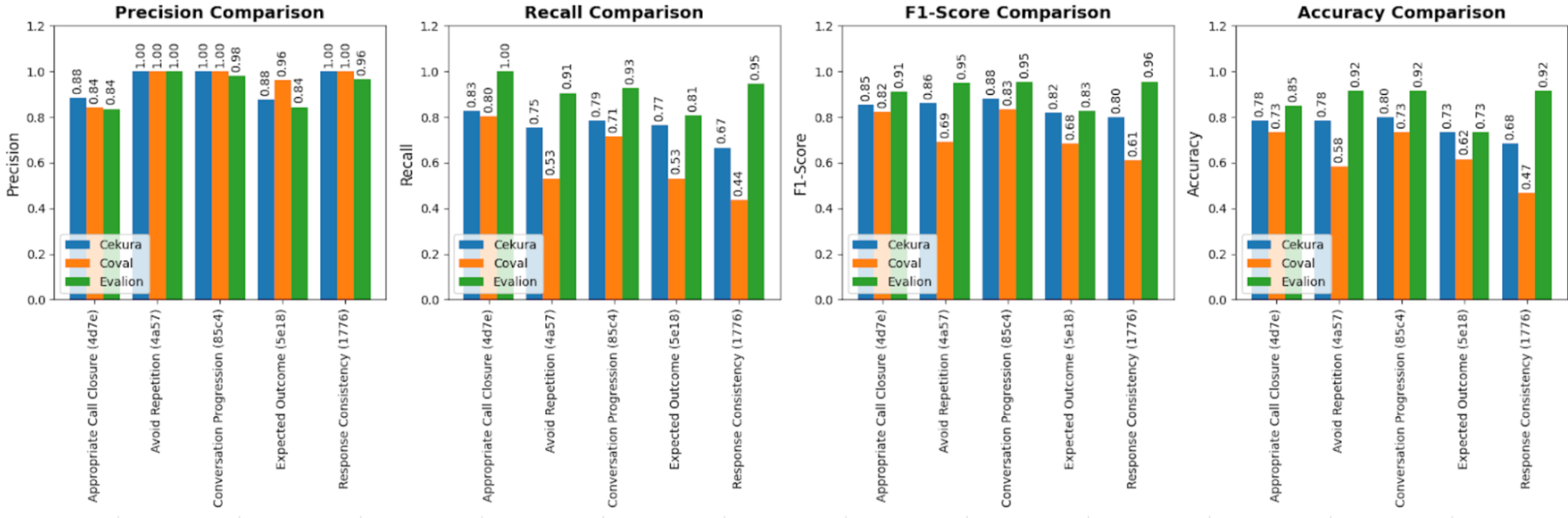}
\caption{A plot of Table~\ref{tab:binary_metrics} for easier viewing and comparison.}
\label{fig:binary_metrics}
\end{figure}

Table~\ref{tab:binary_metrics} presents a comprehensive summary of performance metrics across all platforms and evaluation dimensions. A bar chart counterpart of Table~\ref{tab:binary_metrics} can be found in Figure~\ref{fig:binary_metrics}.

\begin{table*}[t]
\centering
\begin{tabular}{llcccc}
\hline
\textbf{Platform} & \textbf{Metric} & \textbf{Precision} & \textbf{Recall} & \textbf{F1 Score} & \textbf{Accuracy} \\
\hline
\multirow{5}{*}{Evalion} 
& Appropriate Call Closure & 0.836 & \textbf{1.000} & \textbf{0.911} & \textbf{0.850} \\
& Avoid Repetition & \textbf{1.000} & \textbf{0.906} & \textbf{0.950} & \textbf{0.917} \\
& Conversation Progression & 0.981 & \textbf{0.929} & \textbf{0.954} & \textbf{0.917} \\
& Response Consistency & 0.964 & \textbf{0.947} & \textbf{0.956} & \textbf{0.917} \\
& Expected Outcome & 0.844 & \textbf{0.809} & \textbf{0.826} & \textbf{0.733} \\
& \textit{Mean} & \textit{0.925} & \textit{\textbf{0.918}} & \textit{\textbf{0.919}} & \textit{\textbf{0.867}} \\
\hline
\multirow{5}{*}{Cekura} 
& Appropriate Call Closure & 0.884 & 0.826 & 0.854 & 0.783 \\
& Avoid Repetition & \textbf{1.000} & 0.755 & 0.860 & 0.783 \\
& Conversation Progression & \textbf{1.000} & 0.786 & 0.880 & 0.800 \\
& Response Consistency & \textbf{1.000} & 0.667 & 0.800 & 0.683 \\
& Expected Outcome & 0.878 & 0.766 & 0.818 & 0.733 \\
& \textit{Mean} & \textit{\textbf{0.952}} & \textit{0.760} & \textit{0.842} & \textit{0.757} \\
\hline
\multirow{5}{*}{Coval} 
& Appropriate Call Closure & 0.841 & 0.804 & 0.822 & 0.733 \\
& Avoid Repetition & \textbf{1.000} & 0.528 & 0.691 & 0.583 \\
& Conversation Progression & \textbf{1.000} & 0.714 & 0.833 & 0.733 \\
& Response Consistency & \textbf{1.000} & 0.439 & 0.610 & 0.467 \\
& Expected Outcome & \textbf{0.962} & 0.532 & 0.685 & 0.617 \\
& \textit{Mean} & \textit{0.960} & \textit{0.603} & \textit{0.728} & \textit{0.627} \\
\hline
\end{tabular}
\caption{Binary metrics performance across platforms. Bold indicates best performance per metric.}
\label{tab:binary_metrics}
\end{table*}

Results reveal distinct performance patterns across platforms. While all platforms achieved high precision ($>$0.83 minimum), they differ in recall. Evalion shows the most balanced approach with recall rates of 0.809--1.000 and achieves the highest mean F1-score (0.919) with strong performance across all metrics except Expected Outcome (0.826). Cekura demonstrates intermediate performance (mean F1-score: 0.842) with consistent but moderate recall rates, while Coval shows the lowest overall performance (mean F1-score: 0.728) driven primarily by poor recall, as low as 0.439 for Response Consistency.

Metric-specific analysis reveals that Expected Outcome proved universally challenging (F1-scores: 0.826, 0.818, 0.685), likely due to the complexity of assessing task completion. Response Consistency showed the largest performance gap between platforms, with Coval achieving only 0.439 recall compared to Evalion's 0.947. All platforms exhibit Precision $\geq$ Recall for most metrics, indicating a systematic conservative bias, i.e., platforms prefer to withhold positive judgments when uncertain, a pattern most extreme in Coval where perfect precision on three metrics comes at the cost of recall as low as 0.439.

\subsubsection{Continuous Metric Performance}
\label{subsec:eval-continuous-performance}

Customer Satisfaction Score (CSAT) provides a holistic measure of conversation quality on a 1--5 scale. Evaluating continuous metrics presents unique challenges as platforms must predict not just binary outcomes but specific satisfaction levels that align with human perception.

We employ three complementary metrics to evaluate CSAT prediction performance~\cite{chai2014root}:

\begin{itemize}
    \item \textbf{Mean Absolute Error (MAE)} measures the average magnitude of prediction errors: $MAE = \frac{1}{n}\sum_{i=1}^{n}|y_i - \hat{y}_i|$, where $y_i$ represents the human rating and $\hat{y}_i$ the platform prediction. MAE provides an interpretable measure in the same units as the original scale~\cite{willmott2005advantages}. 
    \item \textbf{Root Mean Square Error (RMSE)} penalizes larger errors more heavily: $RMSE = \sqrt{\frac{1}{n}\sum_{i=1}^{n}(y_i - \hat{y}_i)^2}$, making it sensitive to outliers and particularly relevant for identifying platforms prone to severe mispredictions~\cite{hyndman2006another}.
    \item \textbf{Pearson correlation coefficient} measures the linear relationship between predicted and actual scores: $r = \frac{\sum_{i=1}^{n}(y_i - \bar{y})(\hat{y}_i - \bar{\hat{y}})}{\sqrt{\sum_{i=1}^{n}(y_i - \bar{y})^2}\sqrt{\sum_{i=1}^{n}(\hat{y}_i - \bar{\hat{y}})^2}}$, indicating whether platforms correctly capture relative satisfaction differences even if absolute calibration differs~\cite{benesty2009pearson}.
\end{itemize}

These metrics collectively assess different aspects of prediction quality. MAE directly quantifies prediction accuracy in interpretable units (e.g., ``predictions are off by 0.5 scale points on average''), crucial for operational decisions. RMSE identifies platforms that occasionally produce catastrophic errors—a critical consideration for customer service where severe mispredictions could miss escalation opportunities. Correlation reveals whether platforms understand the relative quality ordering of conversations, essential for comparative analytics and agent performance benchmarking~\cite{botchkarev2019performance}.

\begin{figure}[h!]
\centering
\includegraphics[width=\columnwidth]{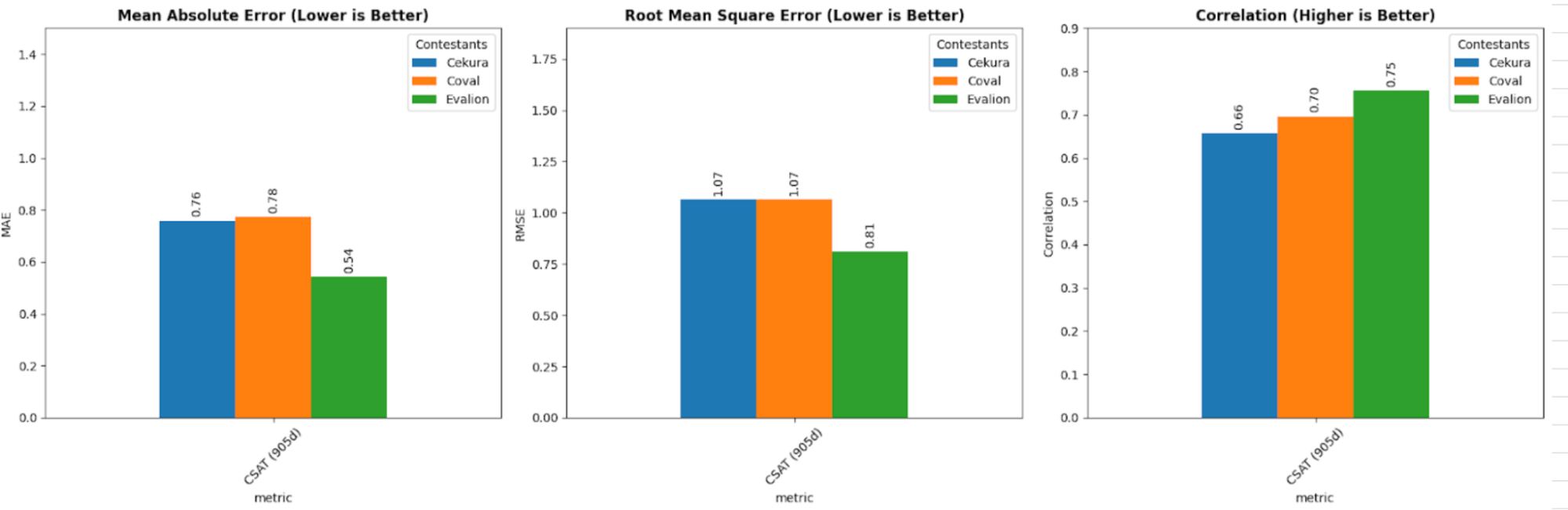}
\caption{A plot of Table~\ref{tab:binary_metrics} for easier viewing and comparison.}
\label{fig:csat_performance}
\end{figure}

Table~\ref{tab:csat_performance} presents CSAT prediction performance across platforms. A bar chart counterpart of Table~\ref{tab:csat_performance} can be found in Figure~\ref{fig:csat_performance}.

\begin{table}[h!]
\centering
\begin{tabular}{lccc}
\hline
\textbf{Platform} & \textbf{MAE} & \textbf{RMSE} & \textbf{Correlation} \\
& (Lower is better) & (Lower is better) & (Higher is better) \\
\hline
Evalion & \textbf{0.542} & \textbf{0.809} & \textbf{0.755} \\
Cekura & 0.758 & 1.067 & 0.658 \\
Coval & 0.775 & 1.067 & 0.697 \\
\hline
\end{tabular}
\caption{CSAT prediction performance across platforms}
\label{tab:csat_performance}
\end{table}

The CSAT results demonstrate notably stronger performance than binary metrics. Evalion achieved exceptional accuracy with MAE of 0.542, meaning predictions deviate by approximately half a scale point on average from human ratings, a 28\% improvement over Cekura (0.758 MAE) and 30\% over Coval (0.775 MAE). The RMSE values show Evalion (0.809) maintains consistently accurate predictions with fewer extreme errors compared to both Cekura and Coval (both 1.067).

Evalion's high correlation (0.755) with human ratings indicates strong alignment with human satisfaction perception, substantially exceeding both Coval (0.697) and Cekura (0.658). A MAE of 0.542 means Evalion's CSAT predictions are typically accurate within half a scale point, approaching human-level inter-rater variability. This accuracy level enables reliable automated satisfaction monitoring for production deployments, with predictions reliable enough for real-time quality monitoring, performance benchmarking across agents or time periods, and automated escalation of low-satisfaction interactions.

\subsubsection{Statistical Significance}

To ensure that the observed performance differences between platforms reflect genuine capability gaps rather than random variation, we conducted statistical significance testing across both binary and continuous evaluation metrics.

\textbf{Binary Metrics}

For binary metrics, we performed a comprehensive statistical analysis using methods specifically designed for paired binary data and bounded classification metrics. Our analysis examines performance across all simulation-metric pairs combined (60 simulations × 5 binary metrics = 300 paired observations per platform), treating each evaluation as an independent observation.

Cochran's Q test \citep{cochran1950comparison} revealed highly significant differences among platforms ($Q(2) = 62.85$, $p < 0.001$), indicating the probability of observing such large differences by chance alone is less than $0.1\%$.

Pairwise comparisons using McNemar's test confirmed all differences remained statistically significant after Bonferroni correction ($\alpha = 0.05/3 = 0.017$). Cohen's kappa ($\kappa$) measures were 0.305 for Evalion vs Cekura, 0.165 for Evalion vs Coval and 0.379 for Cekura vs Coval. For context, Cohen's kappa ($\kappa$) values of 0.0--0.20 indicate slight agreement, 0.21--0.40 fair agreement, and 0.41--0.60 moderate agreement.

For more details on the statistical analysis using these techniques, refer to Appendix~\ref{app:eval-stats}.

\textbf{Continuous Metrics (CSAT)}

To assess the statistical significance of error differences in CSAT prediction performance, we computed the paired t-tests for MAE differences on each pair of providers.

Using paired samples t-tests on absolute errors:
\begin{itemize}
    \item Evalion vs Cekura: $t(59) = -4.23, p < 0.001$
    \item Evalion vs Coval: $t(59) = -4.67, p < 0.001$
    \item Cekura vs Coval: $t(59) = -0.31, p = 0.758$
\end{itemize}

The t-tests on the paired comparison between Evalion against both Coval and Cekura showed that the measured difference in MAE is indeed statistically significant ($p < 0.001$).

\subsubsection{Robustness Validation on F1-Score}

For F1-scores, permutation tests (10,000 iterations) and bootstrap confidence intervals validated that Evalion's superior performance represents genuine capability differences.

\textbf{Bootstrap Confidence Intervals}

First, we constructed 95\% confidence intervals using the percentile bootstrap method with 10,000 resampled iterations. The confidence interval bounds are determined by the 2.5th and 97.5th percentiles of the bootstrap distribution and can be observed in Table~\ref{tab:bootstrap_ci}:

\begin{table}[h]
\centering
\begin{tabular}{|l|c|c|}
\hline
Platform & Mean F1-Score & 95\% Bootstrap CI \\
\hline
Evalion & 0.919 & [0.869, 0.954] \\
Cekura & 0.842 & [0.818, 0.867] \\
Coval & 0.728 & [0.656, 0.801] \\
\hline
\end{tabular}
\caption{Bootstrap confidence intervals for mean F1-scores}
\label{tab:bootstrap_ci}
\end{table}

The non-overlapping confidence intervals between Evalion and Coval indicate statistically different performance.

\textbf{Permutation tests}

Second, to test whether observed differences could occur by chance, we randomly shuffled platform labels 10,000 times (see results in Table~\ref{tab:permutation}):

\begin{table}[h]
\centering
\begin{tabular}{|l|c|c|c|}
\hline
Comparison & F1 Difference & p-value & Significance \\
\hline
Evalion vs Cekura & 0.077 & 0.039 & * \\
Evalion vs Coval & 0.191 & 0.014 & * \\
Cekura vs Coval & 0.114 & 0.052 & n.s. \\
\hline
\end{tabular}
\caption{Permutation test results for F1-score differences}
\label{tab:permutation}
\end{table}

*Significant at $\alpha = 0.05$; n.s. = not significant

For more details on the statistical analysis using these techniques, refer to Appendix~\ref{app:eval-stats}.

\subsubsection{Effect Sizes and Practical Significance}

Beyond statistical significance, we calculated effect sizes to quantify the magnitude of differences:

\begin{table}[h]
\centering
\begin{tabular}{|l|c|c|c|}
\hline
Comparison & Accuracy & Cohen's h & Real-World Impact \\
 & Difference & & (per 1000 evals) \\
\hline
Evalion vs Cekura & 11.0\% & 0.284 (medium) & 110 additional correct \\
Evalion vs Coval & 24.0\% & 0.567 (large) & 240 additional correct \\
Cekura vs Coval & 13.0\% & 0.283 (medium) & 130 additional correct \\
\hline
\end{tabular}
\caption{Effect sizes and real-world impact of platform differences}
\label{tab:effect_sizes}
\end{table}

Cohen's h measures the difference between two proportions using arcsine transformation, where 0.2 = small, 0.5 = medium, and 0.8 = large effect.

\subsubsection{Summary of Evaluation Results}

The evaluation study demonstrates clear performance hierarchy across platforms. Evalion achieves 86.7\% overall accuracy with balanced precision-recall trade-offs, Cekura shows intermediate performance at 75.7\% accuracy with conservative but consistent predictions, and Coval exhibits 62.7\% accuracy limited by extremely conservative recall. These differences are both statistically significant (p $<$ 0.001) and practically meaningful, translating to 110--240 additional correct evaluations per 1000 calls depending on platform choice. The superior CSAT performance, particularly from Evalion (correlation 0.755, MAE 0.542), indicates continuous satisfaction assessment has matured to production-ready accuracy levels, with automated predictions reliable enough to replace or augment human assessment for many use cases.

\section{Conclusion}
This presents the first systematic framework for evaluating the quality of voice AI testing platforms through human-centered benchmarking, addressing a critical gap as the industry scales to billions of daily interactions. Our dual-dimensional evaluation framework---measuring both simulation quality and evaluation accuracy---provides organizations with reproducible methods to assess any testing approach, whether internal tools or commercial platforms.

\subsection{Key Contributions and Findings}

Our empirical validation across three commercial platforms demonstrated the framework's discriminative power, successfully revealing substantial performance variations with significant real-world implications. The framework proved robust enough to capture meaningful differences between platforms, with statistically significant distinctions and clear performance hierarchies emerging across both simulation and evaluation dimensions.

Evalion demonstrated superior performance across both dimensions, achieving a 61.3 simulation quality score [see Table~\ref{tab:sim_overview})] (25\% better than Coval, 42\% better than Cekura) and 86.7\% evaluation accuracy [see Table~\ref{tab:binary_metrics}] (24 percentage points above Coval). These differences are not merely statistical artifacts but translate to meaningful operational impact: for organizations running 1,000 automated tests daily, platform selection determines whether 80--400 defects reach production undetected, and whether 40 hours of analyst time are wasted on false positives.

\subsection{Practical Implications}

For practitioners, our findings provide empirical guidance for platform selection and development. Organizations deploying voice AI in mission-critical applications---healthcare diagnosis, financial transactions, emergency services---now have quantitative evidence for testing platform capabilities. The 86.7\% evaluation accuracy achieved by leading platforms approaches the threshold for production-ready automated testing, though the 13.3\% error rate still requires human oversight for high-stakes applications.
The framework itself, designed to be platform-agnostic and implementation-independent, enables organizations to conduct their own evaluations tailored to specific use cases. By focusing on outcomes rather than technical architecture, it provides a level playing field for comparing various testing approaches, from sophisticated commercial platforms to internal tools built on open-source foundations.

\subsection{A final note on Platform Diversity and Market Evolution}\label{sec:conclusion-personal-note}

While our study revealed performance differences with other leading platforms like Cekura and Coval, it is important to recognize that each platform brings unique strengths and innovations to the rapidly evolving voice AI testing market.

Cekura demonstrated exceptional precision across multiple evaluation metrics, achieving perfect 1.000 precision scores for Avoid Repetition, Conversation Progression, and Response Consistency. This conservative approach—while resulting in lower recall—ensures that when Cekura flags an issue, teams can trust the finding with minimal false positives. For organizations with limited manual review capacity or those operating in high-stakes environments where false alarms carry significant costs, this precision-optimized strategy may be particularly valuable.

Coval's innovative application of autonomous vehicle testing methodologies to conversational AI, as noted by \citep{techcrunch2025coval}, represents a pioneering cross-domain approach that may yield future breakthroughs. The platform showed competitive performance in holistic assessment, achieving a 0.697 CSAT correlation, and demonstrated the highest precision scores across several metrics. Its simulation-first philosophy addresses a different aspect of the testing challenge, potentially offering advantages in scenario diversity and edge case discovery that our evaluation metrics may not fully capture.

The existence of multiple viable platforms with different optimization strategies benefits the entire ecosystem. As voice AI applications diversify across industries—from healthcare to finance to entertainment—different testing requirements will emerge. A platform optimized for catching every possible issue may serve development teams differently than one focused on production monitoring with minimal false alarms. Our framework provides the tools to evaluate these trade-offs, but the "best" platform ultimately depends on an organization's specific needs, risk tolerance, and operational constraints.

Finally, we have seen first hand the velocity at which both platforms are evolving thanks to the talent and passion of their engineering team, so we expect their performance across Simulations and Evaluations quality to evolve, favoring the entire ecosystem.

\subsection{Limitations and Future Work}

A very exciting line of future work involves comparing human-driven simulations to the AI-generated ones described in this paper. A fundamental question is whether humans would rank human generated simulations above AI ones. If they do, we would be able to create a golden set for simulations as well and, from there, measure distance to such ideal simulation for a given scenario and persona (allowing platforms to optimize their simulation engines). Alternatively, if AI based simulations are ranked above human ones, then one could conclude that human based QA is no longer needed in a Voice AI conversational setting. We are making good progress in this direction and we hope to report back soon.

Additional research lines involve extending the proposed framework to multilingual contexts, emotional intelligence assessment, and real-time adaptation capabilities. As voice AI evolves toward more sophisticated reasoning and multi-turn dialogue management, testing frameworks must similarly advance. Integration with automated test generation, adversarial testing methods, and continuous learning systems represents promising directions for maintaining testing quality as voice AI capabilities expand.

Several limitations bound our findings. First, our evaluation focused on customer service interactions with a single production agent; performance may vary across different domains. Second, while we tested 45 meaningful scenario-persona combinations, voice AI's infinite conversational space means edge cases remain unexplored. Third, our human ground truth, though carefully constructed with 10 evaluators and 60 simulations, still contains inherent subjectivity that may not fully capture the complexity of all quality dimensions.

\subsection{Closing Remarks}

As Apple's 2025 Siri delays illustrate, even technology giants struggle to ensure voice AI reliability at scale. This research provides the measurement foundations necessary for the industry to move beyond anecdotal quality assessment toward systematic, reproducible testing validation. By establishing that testing platforms differ significantly in their capabilities---with performance gaps translating to hundreds of missed defects or false alarms daily---we underscore that testing infrastructure deserves the same rigorous evaluation applied to the voice AI systems themselves.

The transition from experimental voice AI demonstrations to trustworthy production systems requires not just better agents, but better methods for validating those agents. Our framework represents a critical step toward that goal, enabling organizations to make evidence-based decisions about their quality assurance infrastructure. As voice AI becomes increasingly central to human-computer interaction, the ability to reliably test these systems becomes not just a technical necessity but an ethical imperative, ensuring that the AI systems we deploy into the world meet the quality standards that billions of users deserve.

\bibliographystyle{plainnat}      
\bibliography{references}

\clearpage
\appendix
\appsection{Simulation Study}{

\subsection{Test Scenarios}
\label{app:test-sceanrios}

Below are the full prompts used in each scenario, together with their  difficulty rating.

\begin{itemize}
    \item Test ID: TC\_001
    \begin{itemize}
        \renewcommand{\labelitemii}{$\circ$}
        \item Prompt: After scheduling a closing you want to find the notary's contact details. Ask where exactly they are displayed.
        \item Difficulty: Easy
    \end{itemize}
    \item Test ID: TC\_002
    \begin{itemize}
        \renewcommand{\labelitemii}{$\circ$}
        \item Prompt: You already activated your account but forgot your password. You want help using the 'Forgot Password' flow, using the temporary password email, and then creating a new password.
        \item Difficulty: Easy
    \end{itemize}
    \item Test ID: TC\_003
    \begin{itemize}
        \renewcommand{\labelitemii}{$\circ$}
        \item Prompt: Inside a loan file you want to locate the banner details, find the Instructions section, and upload additional documents.
        \item Difficulty: Easy
    \end{itemize}
    \item Test ID: TC\_004
    \begin{itemize}
        \renewcommand{\labelitemii}{$\circ$}
        \item Prompt: You're new to the Pipeline View: want to filter for today's closings with pending eSign, sort by Closing Date, and search by address.
        \item Difficulty: Easy
    \end{itemize}
    \item Test ID: TC\_005
    \begin{itemize}
        \renewcommand{\labelitemii}{$\circ$}
        \item Prompt: You need to understand RON scheduling: ask who sets the RON appointment, where to see the Signing Room, and when confirmed details appear.
        \item Difficulty: Medium
    \end{itemize}
    \item Test ID: TC\_006
    \begin{itemize}
        \renewcommand{\labelitemii}{$\circ$}
        \item Prompt: You want to schedule a traditional/hybrid closing using borrower preferences, choose a time slot, set notarization, and confirm if the date can be changed once scheduled.
        \item Difficulty: Medium
    \end{itemize}
    \item Test ID: TC\_007
    \begin{itemize}
        \renewcommand{\labelitemii}{$\circ$}
        \item Prompt: Borrower hasn't finished eSign yet. You uploaded ink-signed docs and saw a prompt. Clarify whether the eSign docs are included or if you need to remind the borrower.
        \item Difficulty: Medium
    \end{itemize}
    \item Test ID: TC\_008
    \begin{itemize}
        \renewcommand{\labelitemii}{$\circ$}
        \item Prompt: You need to activate your account, then immediately schedule a closing, and finally share the closing package with another agent.
        \item Difficulty: Medium
    \end{itemize}
    \item Test ID: TC\_009
    \begin{itemize}
        \renewcommand{\labelitemii}{$\circ$}
        \item You want to share a closing with both an attorney and a notary, but the notary has no account and the attorney already does. Afterward you also want to confirm borrower-preferred scheduling slots.
        \item Difficulty: Medium
    \end{itemize}
    \item Test ID: TC\_010
    \begin{itemize}
        \renewcommand{\labelitemii}{$\circ$}
        \item Prompt: You need to upload a 50MB PDF as a trailing document. Is there a file size limit? While you check that, can you also tell me where in the UI I can see the loan's interest rate? And finally, who gets notified when I upload this trailing document?
        \item Difficulty: Medium
    \end{itemize}
    \item Test ID: TC\_011
    \begin{itemize}
        \renewcommand{\labelitemii}{$\circ$}
        \item Prompt: You are a notary who has just been sent a closing package by a settlement agent. You need to activate your Notary Workspace account, choose the MFA method, complete your profile, and access documents.
        \item Difficulty: Hard
    \end{itemize}
    \item Test ID: TC\_012
    \begin{itemize}
        \renewcommand{\labelitemii}{$\circ$}
        \item Prompt: You must upload ink-signed docs, handle the system prompt that appears because the borrower’s eSign is still incomplete, and then find notary contact info to remind them.
        \item Difficulty: Hard
    \end{itemize}
    \item Test ID: TC\_013
    \begin{itemize}
        \renewcommand{\labelitemii}{$\circ$}
        \item Prompt: You have a list of 4 very specific feature requests, framed as questions. 1. "How do I set up a rule to auto-assign all loans from 'Lender X' to my 'High-Value' team?" 2. "Where is the audit log to see which team member downloaded a closing package?" 3. "How do I generate a report of all my closings from last quarter?" 4. "How do I reset my password?".
        \item Difficulty: Hard
    \end{itemize}
    \item Test ID: TC\_014
    \begin{itemize}
        \renewcommand{\labelitemii}{$\circ$}
        \item Prompt: You have a bad connection that cuts out occasionally. You need to reset your password, but the email isn't arriving. You also need to upload a document for a closing that is happening in 10 minutes, but the upload button isn't working. Mid-call, you must interrupt the agent and ask if B***d offers services in Canada.
        \item Difficulty: Hard
    \end{itemize}
    \item Test ID: TC\_015
    \begin{itemize}
        \renewcommand{\labelitemii}{$\circ$}
        \item Prompt: You are a settlement agent. You shared a closing with a notary. The notary claims they completed their profile but you still can't see their phone number in the Contacts tab. You also need to know if you can set a default notary for all your closings.
        \item Difficulty: Hard
    \end{itemize}
\end{itemize}

\subsection{Tested Personas}
\label{app:personas-prompt}

Below are the full prompts used for each persona.

\textbf{Olivia}

Impatient, easily frustrated, and has a short temper. She feels wronged and is quick to express her anger and dissatisfaction.
Her motivation is to have her problem solved immediately and to feel heard and validated in her anger.
She believes being aggressive is the only way to get results.
- Key Characteristics: She may use sarcasm, interrupt the agent, and threaten to escalate the issue or leave a negative review.
- Communication Style: Blunt, demanding, and often emotional. Uses phrases like "This is unacceptable," "I want to speak to a manager," and "You're not listening to me."

\textbf{Sophia}

Always short on time and values efficiency above all else.
She is likely multitasking during the conversation (e.g., checking emails, in transit).
She is direct, concise, and wants to get to the point as quickly as possible.
Her motivation is to resolve the issue with minimal time and effort. Appreciates clear, actionable information and dislikes small talk or lengthy explanations.
- Communication Style: To the point and professional. Uses short sentences and expects the same in return. May say things like, "I only have a few minutes," "What's the bottom line?" or "Just tell me what I need to do."

\textbf{Chloe}

Reasonable, calm, and generally cooperative.
She is looking for a fair and straightforward resolution.
While she may be initially disappointed or concerned, she is willing to work with the agent to find a solution.
Her motivation is to understand the situation, find a reasonable solution, and have a pleasant interaction.
- Communication Style: Polite, clear, and patient. Asks questions to clarify information and is generally understanding of the process.

\subsection{Scoring variants results}
\label{app:socring-variants-res}


\begin{table}[h!]
\centering
\small
\adjustbox{width=\textwidth}{
\begin{tabular}{|l|c|c|c|c|c|c|c|c|}
\hline
\textbf{Contestant} & \textbf{League - WD} & \textbf{League - ND} & \textbf{League - WD - PCA} & \textbf{League - ND - PCA} & \textbf{Elo - WD} & \textbf{Elo - ND} & \textbf{Elo - WD - PCA} & \textbf{Elo - ND - PCA} \\
\hline
Cekura & 43.04 & 30.39 & 43.95 & 32.21 & 44.71 & 43.36 & 45.44 & 44.36 \\
\hline
Coval & 48.92 & 35.40 & 52.15 & 39.80 & 50.00 & 48.13 & 51.76 & 52.19 \\
\hline
Evalion & 61.31 & 47.66 & 64.80 & 52.56 & 60.51 & 62.75 & 64.15 & 66.49 \\
\hline
\end{tabular}
}
\caption{Overall Score Results}
\end{table}

\begin{table}[h!]
\centering
\small
\adjustbox{width=\textwidth}{
\begin{tabular}{|l|c|c|c|c|c|c|c|c|}
\hline
\textbf{Contestant} & \textbf{League - WD} & \textbf{League - ND} & \textbf{League - WD - PCA} & \textbf{League - ND - PCA} & \textbf{Elo - WD} & \textbf{Elo - ND} & \textbf{Elo - WD - PCA} & \textbf{Elo - ND - PCA} \\
\hline
Cekura & 37.17 & 25.44 & 39.26 & 27.07 & 39.45 & 37.00 & 41.51 & 38.92 \\
\hline
Coval & 49.11 & 36.44 & 53.36 & 40.98 & 51.30 & 48.32 & 52.53 & 53.83 \\
\hline
Evalion & 63.72 & 51.11 & 68.29 & 56.52 & 61.83 & 65.17 & 67.31 & 70.29 \\
\hline
\end{tabular}
}
\caption{Scenario Adherence Results}
\end{table}

\begin{table}[h!]
\centering
\small
\adjustbox{width=\textwidth}{
\begin{tabular}{|l|c|c|c|c|c|c|c|c|}
\hline
\textbf{Contestant} & \textbf{League - WD} & \textbf{League - ND} & \textbf{League - WD - PCA} & \textbf{League - ND - PCA} & \textbf{Elo - WD} & \textbf{Elo - ND} & \textbf{Elo - WD - PCA} & \textbf{Elo - ND - PCA} \\
\hline
Cekura & 41.11 & 25.56 & 41.36 & 29.47 & 42.80 & 40.67 & 42.50 & 41.41 \\
\hline
Coval & 46.78 & 30.78 & 51.87 & 39.40 & 47.79 & 45.95 & 52.03 & 52.15 \\
\hline
Evalion & 62.11 & 46.56 & 67.67 & 55.70 & 61.99 & 63.86 & 66.83 & 69.48 \\
\hline
\end{tabular}
}
\caption{Human Naturalness Results}
\end{table}

\begin{table}[h!]
\centering
\small
\adjustbox{width=\textwidth}{
\begin{tabular}{|l|c|c|c|c|c|c|c|c|}
\hline
\textbf{Contestant} & \textbf{League - WD} & \textbf{League - ND} & \textbf{League - WD - PCA} & \textbf{League - ND - PCA} & \textbf{Elo - WD} & \textbf{Elo - ND} & \textbf{Elo - WD - PCA} & \textbf{Elo - ND - PCA} \\
\hline
Cekura & 52.79 & 41.81 & 52.79 & 41.81 & 53.62 & 54.55 & 53.62 & 54.55 \\
\hline
Coval & 50.82 & 38.62 & 50.82 & 38.62 & 50.48 & 50.06 & 50.48 & 50.06 \\
\hline
Evalion & 57.29 & 44.15 & 57.29 & 44.15 & 57.26 & 58.42 & 57.26 & 58.42 \\
\hline
\end{tabular}
}
\caption{Persona Adherence Results}
\end{table}

The full set of metrics scores, for each variant, scenario, persona and provider, are reported in the spreadhseet \\ \url{https://docs.google.com/spreadsheets/d/1FHw0pPdlX2d32XvL1TIbhYGbXL2VZkVuih5oP3BI3_Y}.

See Section~\ref{app:code-and-files} for the code to generate this data.

\subsection{Code, transcripts and survey files}
\label{app:code-and-files}

The code to reproduce the analysis, as well as the survey and response files, are all available in the accompanying repository \url{https://github.com/evalionai/research}. The conversational transcripts and audio files surveyed in this study are proprietary to Sei \citep{sei} and we are not authorised to make them publicly available. For confidentiality reasons, only obfuscated versions of the transcripts are included in the repository. Access to the original transcripts and audio files may be granted for research purposes upon reasonable request, subject to client approval and confidentiality agreements. All methodological details are reported in the paper to ensure reproducibility. Researchers interested in data access may contact \href{mailto:research@evalion.ai}{research@evalion.ai}.

\subsection{Regression Analysis on Overall Scenario Adherence and Overall Human Naturalness metrics}
\label{app:simulation_regression}

\begin{figure}[H]
\centering
\includegraphics[width=0.7\columnwidth]{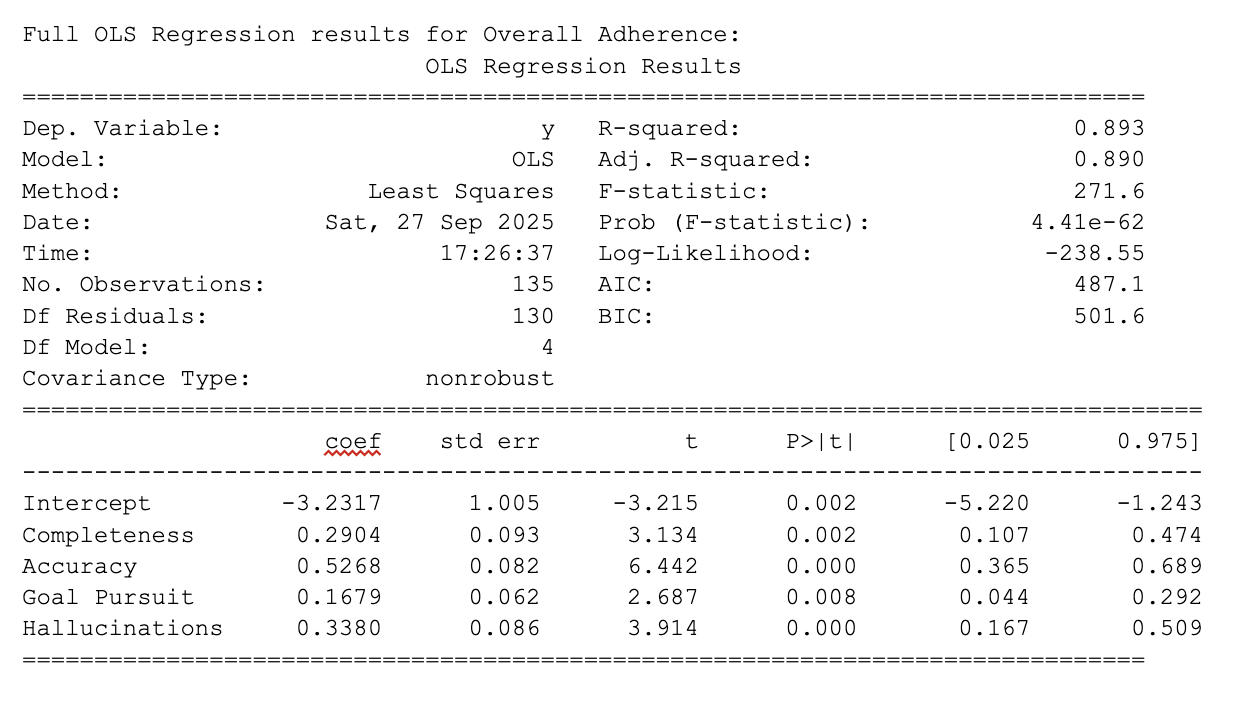}
\caption{OLS regression analysis of Overall Scenario Adherence metric.}
\label{fig:OLS-Scenario-Adherence}
\end{figure}

\begin{figure}[H]
\centering
\includegraphics[width=0.7\columnwidth]{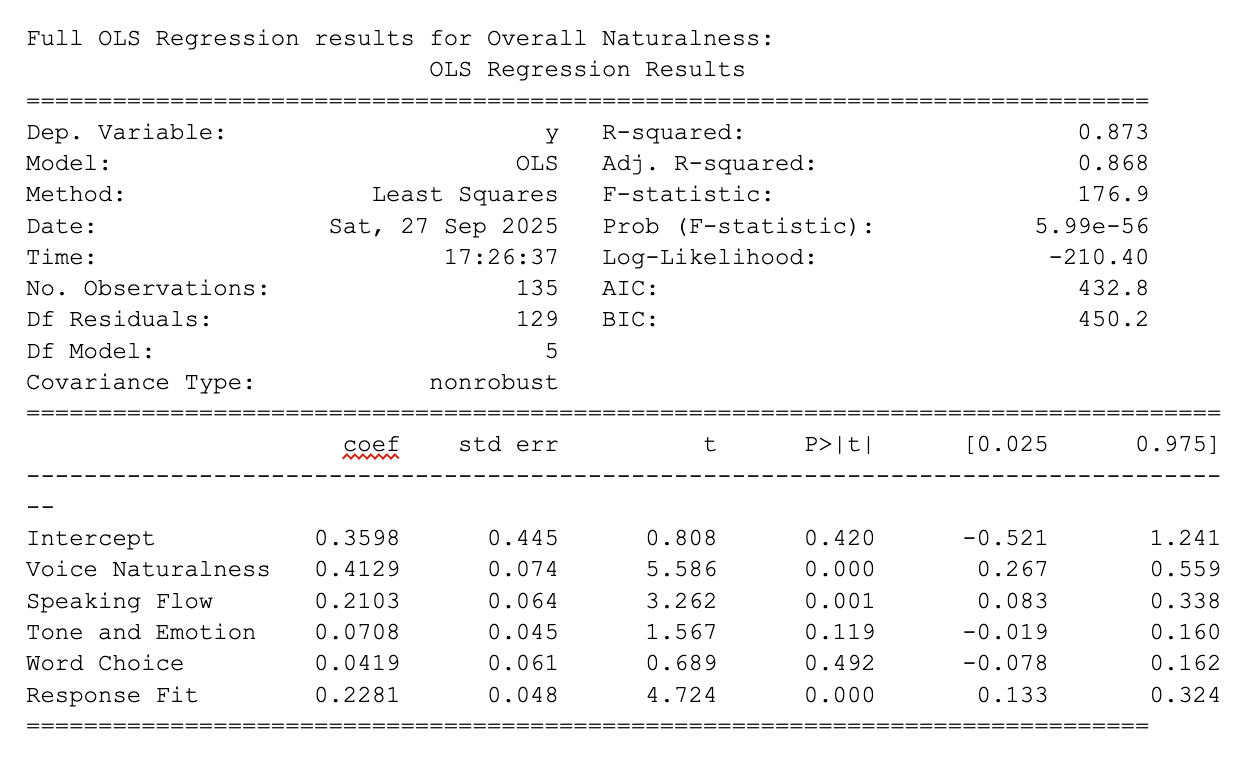}
\caption{OLS regression analysis of Overall Human Naturalness metric.}
\label{fig:OLS-Overall-Human}
\end{figure}

}
\appsection{Evaluation Study Details}{

\subsection{Custom Evaluation Metrics}
\label{app:eval-prompts}

When metrics were not natively available in a platform, we implemented them as custom metrics using the following standardized definitions:

\paragraph{CSAT (Customer Satisfaction Score)}
\begin{quote}
Given the transcript, how satisfied would you be with the customer support service, if you were the customer on that call?

Assume that your role in the transcript is ``\{\{user\}\}''.

Give a score between 1 and 5, where 1 is ``very unsatisfied'' and 5 is ``very satisfied''.
\end{quote}

\paragraph{Appropriate Call Closure}
\begin{quote}
Given the transcript, did the \{\{assistant\}\} avoid ending the call abruptly mid-conversation without proper closing statements?

Return YES if:
\begin{itemize}
\item The \{\{assistant\}\}'s final message contains explicit closing language (e.g., ``Goodbye,'' ``Have a great day,'' ``Thank you for calling'')
\item OR the \{\{assistant\}\}'s final message indicates a transfer with appropriate handoff language
\item OR the \{\{assistant\}\}'s final message clearly signals conversation completion (e.g., ``That covers everything,'' ``Those are all the details'')
\end{itemize}

Return NO if:
\begin{itemize}
\item The \{\{assistant\}\}'s final message is ongoing conversation (question, explanation, request for information, etc.)
\item OR the \{\{assistant\}\}'s final message ends mid-topic without closure or resolution
\item OR the conversation appears cut off during active discussion
\item OR the \{\{assistant\}\}'s last message expects a user response but the conversation ends
\end{itemize}
\end{quote}

\paragraph{Avoid Repetition}
\begin{quote}
Given the transcript, did the \{\{assistant\}\} avoid repeating the same response or entering a conversational loop?

Return YES if (all of the points):
\begin{itemize}
\item The \{\{assistant\}\} provides distinct, forward-moving responses throughout the interaction
\item AND there is no repetition of the same sentence, phrase, or instruction multiple times without \{\{user\}\} prompting
\item AND the \{\{assistant\}\} does not return to the same point or response structure repeatedly (e.g., circular explanations or looped fallback phrases)
\end{itemize}

Return NO if:
\begin{itemize}
\item The \{\{assistant\}\} repeats the same response or phrase multiple times unnecessarily
\item OR the \{\{assistant\}\} enters a conversational loop, revisiting the same statements or fallback messages without progressing the conversation
\end{itemize}
\end{quote}

\paragraph{Conversation Progression}
\begin{quote}
Given the transcript, did the \{\{assistant\}\} effectively advance the conversation toward resolution?

Return YES if:
\begin{itemize}
\item Each \{\{assistant\}\}'s response moved the conversation closer to addressing the \{\{user\}\}'s need
\item OR the \{\{assistant\}\} proactively requested necessary information to resolve the query
\end{itemize}

Return NO ONLY if:
\begin{itemize}
\item The \{\{assistant\}\}'s responses were circular or redundant
\item OR the \{\{assistant\}\} failed to ask for critical information needed to address the query
\item OR the conversation stalled due to the \{\{assistant\}\}'s inability to progress
\end{itemize}
\end{quote}

\paragraph{Response Consistency}
\begin{quote}
Given the transcript, evaluate whether the \{\{assistant\}\} responded consistently across the duration of the call.

Return YES if (all of the points):
\begin{itemize}
\item The \{\{assistant\}\} maintained consistent tone and communication style throughout
\item AND information provided remained factually consistent with no contradictions about policies, procedures, or facts
\item AND similar questions or requests were handled in a comparable manner
\end{itemize}

Return NO ONLY if:
\begin{itemize}
\item The \{\{assistant\}\} provided contradictory information or forgot previous statements made in the same call
\item OR there were significant shifts in tone or engagement without clear justification
\item OR similar requests were handled very differently or inconsistently
\end{itemize}
\end{quote}

\paragraph{Expected Outcome}
\begin{quote}
Given the transcript, did the conversation reach the following expected outcome?

Note that the expected outcome is a bullet-point list of statements.\\
Return YES, if all of the statements are true for the conversation.\\
Return NO, if any of the statements is false.

Note also that ``agent'' in the expected outcome refers to ``\{\{assistant\}\}'' in the transcript, and ``user'' in the expected outcome refers to ``\{\{user\}\}'' in the transcript.

Expected outcome:\\
\{\{expected\_outcome\}\}
\end{quote}

\subsection{Human Evaluation Questions}
\label{app:human-questions}
For the human evaluation study, we translated the technical metric definitions into accessible questions for non-expert evaluators:

\paragraph{CSAT}
``How satisfied would you be with the customer support service, if you were the customer (user) on that call? Give a score between 1 and 5, where 1 is `very unsatisfied' and 5 is `very satisfied'.''

\paragraph{Appropriate Call Closure}
``Did the assistant appropriately end the call? Examples of not ending a call appropriately could be ending the call mid-conversation or without a goodbye.''

\paragraph{Avoid Repetition}
``Did the assistant manage to avoid repeating the same response or entering a conversational loop?''

\paragraph{Conversation Progression}
``Did the assistant help move the conversation toward solving the problem? Each assistant's response should bring the conversation closer to resolving the reason for the call.''

\paragraph{Response Consistency}
``Did the assistant stay consistent throughout the conversation? That is, did the assistant keep the same tone of voice, gave answers that didn't contradict each other, and handled similar questions in the same way? For example, not saying yes to a refund early on and then no later without good reason.''

\paragraph{Expected Outcome}
``Did the conversation reach the following expected outcome? Note that the expected outcome is a bullet-point list of statements. Answer YES, if all of the statements are true for the conversation. Answer NO, if any of the statements is false. Expected outcome: \{\{expected\_outcome\}\}''

\subsection{Results with Filtered Dataset (45 Recordings)}
\label{app:filtered-res}

This section presents results after removing 15 recordings with weak consensus on 2 or more metrics, retaining only the 45 recordings with stronger human agreement. This filtered analysis serves as a robustness check for our main findings.

\paragraph{Ground Truth Data Distribution}
Figure~\ref{fig:ground_truth_filtered} shows the distribution of ``yes" answers when we filter out recordings which have any disagreement, and Figure ~\ref{fig:csat_filtered} shows the corresponding CSAT distribution on that set.

\begin{figure}[h!]
\centering
\includegraphics[width=0.5\textwidth]{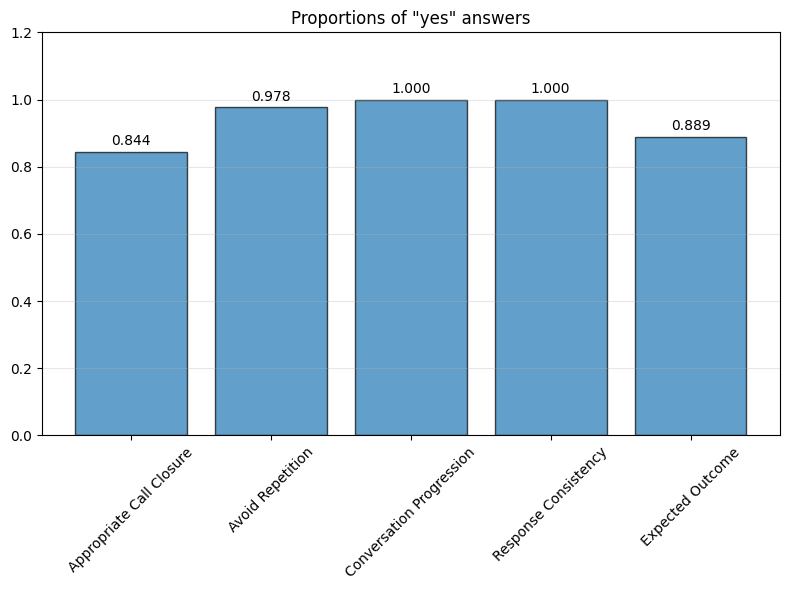}
\caption{Distribution of positive evaluations for binary metrics in the filtered dataset (45 recordings).}
\label{fig:ground_truth_filtered}
\end{figure}

\begin{figure}[h!]
\centering
\includegraphics[width=0.5\textwidth]{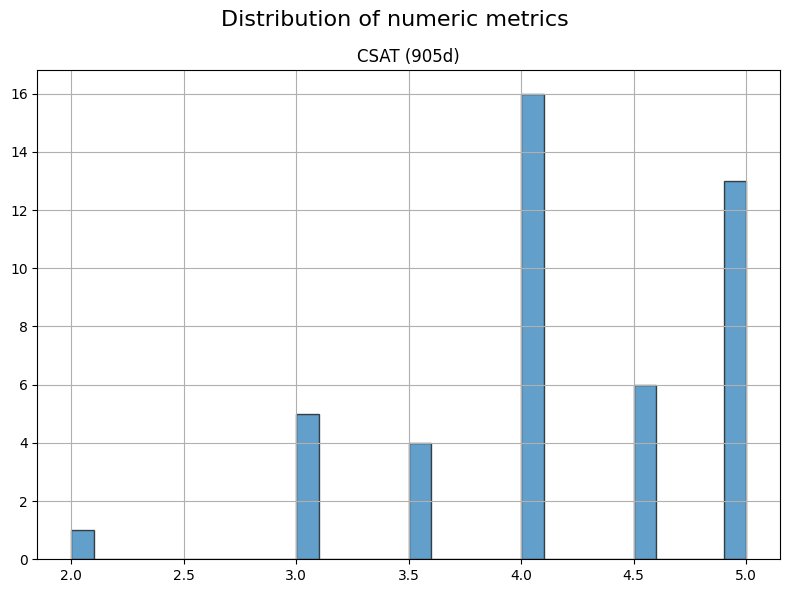}
\caption{CSAT distribution in the filtered dataset (45 recordings).}
\label{fig:csat_filtered}
\end{figure}

\paragraph{Binary Metrics Performance}
Table~\ref{tab:binary_filtered} shows the binary metrics performance obtained with the filtered dataset (45 recordings) and Figure~\ref{fig:binary_filtered} shows a bar chart of the data in Table~\ref{tab:binary_filtered}

\begin{table}[h!]
\centering
\begin{tabular}{llcccc}
\hline
\textbf{Platform} & \textbf{Metric} & \textbf{Precision} & \textbf{Recall} & \textbf{F1 Score} & \textbf{Accuracy} \\
\hline
\multirow{5}{*}{Evalion}
& Appropriate Call Closure & 0.884 & 1.000 & 0.938 & 0.889 \\
& Avoid Repetition & 1.000 & 0.909 & 0.952 & 0.911 \\
& Conversation Progression & 1.000 & 0.956 & 0.977 & 0.956 \\
& Response Consistency & 1.000 & 0.933 & 0.966 & 0.933 \\
& Expected Outcome & 0.921 & 0.875 & 0.897 & 0.822 \\
\hline
\multirow{5}{*}{Cekura}
& Appropriate Call Closure & 0.943 & 0.868 & 0.904 & 0.844 \\
& Avoid Repetition & 1.000 & 0.795 & 0.886 & 0.800 \\
& Conversation Progression & 1.000 & 0.822 & 0.902 & 0.822 \\
& Response Consistency & 1.000 & 0.689 & 0.816 & 0.689 \\
& Expected Outcome & 0.917 & 0.825 & 0.868 & 0.778 \\
\hline
\multirow{5}{*}{Coval}
& Appropriate Call Closure & 0.892 & 0.868 & 0.880 & 0.800 \\
& Avoid Repetition & 1.000 & 0.568 & 0.725 & 0.578 \\
& Conversation Progression & 1.000 & 0.800 & 0.889 & 0.800 \\
& Response Consistency & 1.000 & 0.511 & 0.676 & 0.511 \\
& Expected Outcome & 1.000 & 0.600 & 0.750 & 0.644 \\
\hline
\end{tabular}
\caption{Binary metrics performance on filtered dataset (45 recordings)}
\label{tab:binary_filtered}
\end{table}

\begin{figure}[h!]
\centering
\includegraphics[width=\textwidth]{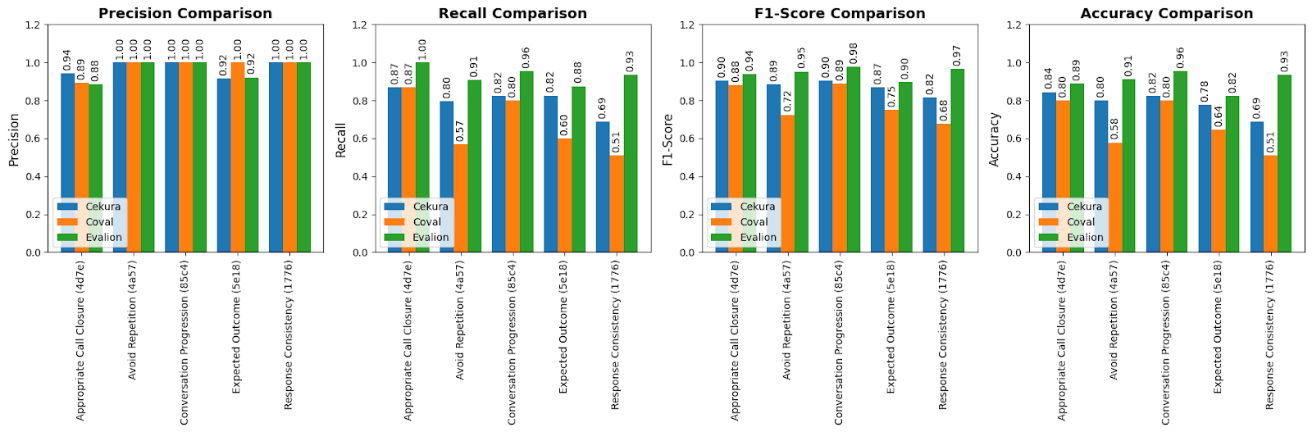}
\caption{Bar chart showing the data in Table~\ref{tab:binary_filtered}}
\label{fig:binary_filtered}
\end{figure}

\paragraph{Continuous Metric Performance}

\begin{table}[h!]
\centering
\begin{tabular}{lccc}
\hline
\textbf{Platform} & \textbf{MAE} & \textbf{RMSE} & \textbf{Correlation} \\
\hline
Evalion & 0.578 & 0.850 & 0.634 \\
Cekura & 0.778 & 1.111 & 0.532 \\
Coval & 0.778 & 1.070 & 0.574 \\
\hline
\end{tabular}
\caption{CSAT prediction performance on filtered dataset (45 recordings)}
\label{tab:csat_filtered}
\end{table}

\begin{figure}[h!]
\centering
\includegraphics[width=\textwidth]{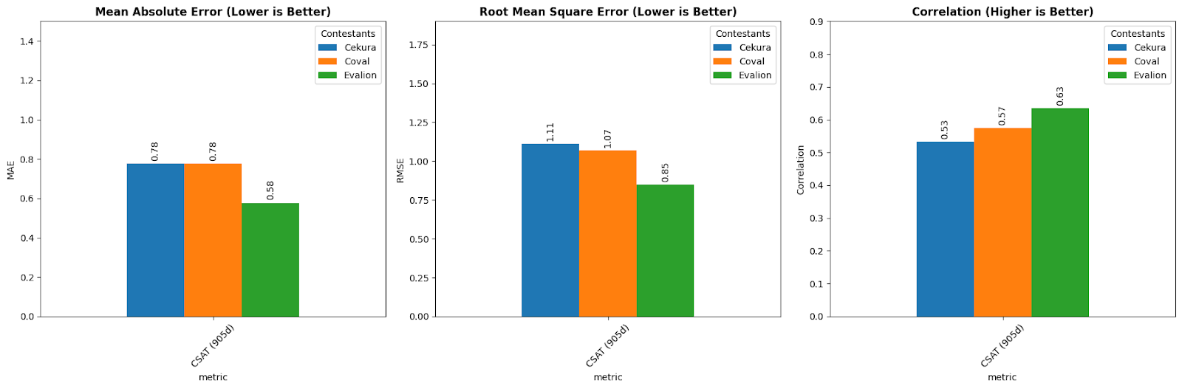}
\caption{Bar chart showing the data in Table~\ref{tab:csat_filtered}.}
\label{fig:binary_filtered-csat}
\end{figure}

\paragraph{Comparison with Full Dataset}

The filtered dataset results show similar platform rankings but with slightly improved absolute performance across all platforms. Evalion maintains its leading position with marginally higher accuracy (89.1\% vs. 86.7\% in full dataset), while the relative performance gaps between platforms remain consistent. This confirms that our decision to include all 60 recordings provides a more conservative and realistic assessment of real-world platform performance, where ambiguous cases are inevitable.

}
\appsection{Statistical Analysis of Evaluation Results}{
\label{app:eval-stats}

\subsection{Statistical Methods Overview}

To determine whether observed performance differences between platforms represent genuine capability differences rather than random variation, we employed multiple complementary statistical approaches specifically designed for our data structure: paired binary observations and bounded classification metrics.

\subsection{Binary Metrics Statistical Analysis}

\subsubsection{Data Structure and Test Selection}

Our analysis examined 300 paired binary observations per platform (60 recordings $\times$ 5 binary metrics), where each observation represented whether a platform's evaluation matched the human ground truth (1 = correct, 0 = incorrect). This paired structure is crucial---all three platforms evaluated the exact same recordings, making the observations dependent rather than independent.

We employed Cochran's Q test as our primary omnibus test to assess whether all three platforms have equal accuracy. This test extends McNemar's test to more than two related samples and is specifically designed for comparing multiple classifiers on the same dataset. For pairwise platform comparisons, we used McNemar's test, which examines the discordant pairs (cases where one platform succeeds and another fails on the same recording-metric combination).

\subsubsection{Overall Performance Differences}

The Cochran's Q test revealed highly significant differences among platforms:
\begin{itemize}
    \item \textbf{Q(2) = 62.85, p $<$ 0.001}
\end{itemize}

This indicates that the probability of observing such large differences between platforms by chance alone is less than 0.1\%, providing strong evidence that the platforms genuinely differ in their evaluation accuracy.

\subsubsection{Pairwise Platform Comparisons}

With Bonferroni correction for multiple comparisons ($\alpha = 0.05/3 = 0.017$), all pairwise comparisons remained statistically significant:

\begin{table}[h]
\centering
\begin{tabular}{|l|c|c|c|}
\hline
Comparison & McNemar & p-value & Cohen's $\kappa$ \\
 & Statistic & (corrected) & \\
\hline
Evalion vs Cekura & 16.00 & 0.0002*** & 0.305 \\
Evalion vs Coval & 15.00 & $<$0.0001*** & 0.165 \\
Cekura vs Coval & 21.00 & 0.0001*** & 0.379 \\
\hline
\end{tabular}
\caption{Pairwise platform comparisons using McNemar's test}
\label{tab:mcnemar-app}
\end{table}

The test uses exact computation when feasible, falling back to the chi-square approximation when necessary. The contingency table format for McNemar's test is:
\[
\begin{bmatrix}
n_{11} & n_{10} \\
n_{01} & n_{00}
\end{bmatrix}
\]
where $n_{11}$ represents both platforms correct, $n_{10}$ represents only platform 1 correct, $n_{01}$ represents only platform 2 correct, and $n_{00}$ represents both incorrect.

Cohen's kappa ($\kappa$) measures agreement beyond chance, where values of 0.0--0.20 indicate slight agreement, 0.21--0.40 fair agreement, and 0.41--0.60 moderate agreement.

\subsubsection{Metric-Level Performance Analysis}

Individual metric analysis revealed varying platform performance across evaluation dimensions:

\begin{table}[h!]
\centering
\begin{tabular}{lccc}
\hline
\textbf{Metric} & \textbf{$\chi^2$} & \textbf{p-value} & \textbf{Max Difference} \\
\hline
Appropriate Call Closure & 2.47 & 0.291 & 11.7\% \\
Avoid Repetition & 18.58 & $<$0.001*** & 33.3\% \\
Conversation Progression & 6.90 & 0.032* & 18.3\% \\
Response Consistency & 28.36 & $<$0.001*** & 45.0\% \\
Expected Outcome & 2.57 & 0.277 & 11.7\% \\
\hline
\end{tabular}
\caption{Chi-square tests for individual metrics}
\end{table}

\subsection{F1-Score Statistical Analysis}

\subsubsection{Bootstrap Confidence Intervals}

We construct 95\% confidence intervals using the percentile bootstrap method with 10,000 resampled iterations. The confidence interval bounds are determined by the 2.5th and 97.5th percentiles of the bootstrap distribution:

\begin{table}[h]
\centering
\begin{tabular}{|l|c|c|}
\hline
Platform & Mean F1-Score & 95\% Bootstrap CI \\
\hline
Evalion & 0.919 & [0.869, 0.954] \\
Cekura & 0.842 & [0.818, 0.867] \\
Coval & 0.728 & [0.656, 0.801] \\
\hline
\end{tabular}
\caption{Bootstrap confidence intervals for mean F1-scores}
\label{tab:bootstrap_ci-app}
\end{table}

The non-overlapping confidence intervals between Evalion and Coval indicate statistically different performance.

\subsubsection{Permutation Tests}

To test whether observed differences could occur by chance, we randomly shuffled platform labels 10,000 times:

\begin{table}[h]
\centering
\begin{tabular}{|l|c|c|c|}
\hline
Comparison & F1 Difference & p-value & Significance \\
\hline
Evalion vs Cekura & 0.077 & 0.039 & * \\
Evalion vs Coval & 0.191 & 0.014 & * \\
Cekura vs Coval & 0.114 & 0.052 & n.s. \\
\hline
\end{tabular}
\caption{Permutation test results for F1-score differences}
\label{tab:permutation-app}
\end{table}

The empirical p-value is calculated as:
\[
p_{perm} = \frac{|\{\Delta^*_i : |\Delta^*_i| \geq |\Delta_{obs}|\}|}{N_{perm}}
\]
where $\Delta_{obs}$ is the observed difference and $\Delta^*_i$ is the difference under permutation $i$.

*Significant at $\alpha = 0.05$; n.s. = not significant

\subsubsection{Performance Consistency Analysis}

Coefficient of Variation (CV) measures relative variability:
\[
CV = \frac{\sigma}{\mu} \times 100\%
\]

\begin{table}[h]
\centering
\begin{tabular}{|l|c|c|c|}
\hline
Platform & Mean F1 & CV & F1 Range \\
\hline
Evalion & 0.919 & 5.4\% & 0.826--0.956 \\
Cekura & 0.842 & 3.5\% & 0.800--0.880 \\
Coval & 0.728 & 11.8\% & 0.610--0.833 \\
\hline
\end{tabular}
\caption{Performance consistency across platforms}
\label{tab:consistency}
\end{table}

Lower CV values indicate more consistent performance across metrics. Cekura shows the most consistent performance (CV = 3.5\%), followed by Evalion (5.4\%), while Coval exhibits greater variability (11.8\%) in F1-scores across different evaluation dimensions.

\subsection{CSAT Performance Statistical Analysis}

For the continuous CSAT metric, we employed different statistical approaches:

\subsubsection{Correlation Analysis}

Pearson correlation coefficients with 95\% confidence intervals:
\begin{itemize}
    \item Evalion: $r = 0.755$ [0.701, 0.802]
    \item Coval: $r = 0.697$ [0.635, 0.750]
    \item Cekura: $r = 0.658$ [0.589, 0.718]
\end{itemize}

\subsubsection{Paired t-tests for MAE Differences}

Using paired samples t-tests on absolute errors:
\begin{itemize}
    \item Evalion vs Cekura: $t(59) = -4.23, p < 0.001$
    \item Evalion vs Coval: $t(59) = -4.67, p < 0.001$
    \item Cekura vs Coval: $t(59) = -0.31, p = 0.758$
\end{itemize}

}

\end{document}